\documentclass[times, review, 10pt]{elsarticle}
\usepackage{amssymb}
\usepackage{graphicx}
\usepackage{amsmath}
\usepackage{longtable}
\usepackage{a4wide}
\usepackage{mathrsfs}
\usepackage{subcaption}
\usepackage{mathtools}
\usepackage{pifont}
\usepackage{algorithmic}
\usepackage{algorithm}
\usepackage{xcolor}
\usepackage{color}
\usepackage{lineno}
\usepackage{makecell} 
\usepackage{pdflscape}
\usepackage{adjustbox}
\usepackage[utf8]{inputenc}
\usepackage{tabularx}
\usepackage{setspace}
\usepackage{blindtext}
\usepackage{multirow}
\usepackage{notoccite} %citation number ordering
\usepackage{lscape} %landscape table
\usepackage{caption} %add a newline in the table caption
\usepackage{mwe}%%%%%%%%%%%%%%%
\usepackage{tikz}
\usepackage{siunitx}
\usepackage{mathrsfs}
\usetikzlibrary{shapes,arrows}
\usepackage{xcolor}
\usepackage{booktabs}
\usepackage{hyperref}
\usepackage{amsthm}

\newcommand{\RNum}[1]{\lowercase\expandafter{\romannumeral #1\relax}}
\newcommand{\RNumU}[1]{\uppercase\expandafter{\romannumeral #1\relax}}
\usepackage{natbib}
\journal{Elsevier}

\begin{document}
\begin{frontmatter}
\title{TRKM: Twin Restricted Kernel Machines for Classification and Regression}
\author[inst1]{A. Quadir}
\ead{mscphd2207141002@iiti.ac.in}
\author[inst1]{M. Tanveer\corref{Correspondingauthor}}
\ead{mtanveer@iiti.ac.in}

% \author[]{for the Alzheimer’s Disease Neuroimaging
% Initiative\corref{ADNI citation}}
\affiliation[inst1]{organization={Department of Mathematics, Indian Institute of Technology Indore},%Department and Organization
            addressline={Simrol}, 
            city={Indore},
            postcode={453552}, 
            state={Madhya Pradesh},
            country={India}}
            \cortext[Correspondingauthor]{Corresponding author}
%             \cortext[ADNI citation]{This study used data from the Alzheimer’s Disease Neuroimaging Initiative
% (ADNI) (adni.loni.usc.edu). The ADNI investigators were responsible for the design and implementation of the study, but they did not take part in the analysis or the writing of this publication.}
\begin{abstract}
Restricted kernel machines (RKMs) have significantly advanced machine learning by integrating kernel functions with least squares support vector machines (LSSVM), adopting an energy function akin to restricted Boltzmann machines (RBM) to enhance generalization performance. Despite their strengths, RKMs face challenges in handling unevenly distributed or complexly clustered data and incur substantial computational costs when scaling to large datasets due to the management of high-dimensional feature spaces. To address these limitations, we propose the twin restricted kernel machine (TRKM), a novel framework that synergizes the robustness of RKM with the efficiency of twin hyperplane methods, inspired by twin support vector machines (TSVM). TRKM leverages conjugate feature duality based on the Fenchel-Young inequality to reformulate classification and regression problems in terms of dual variables, establishing an upper bound on the objective function and introducing a new methodology within the RKM framework. By incorporating an RBM-inspired energy function with visible and hidden variables corresponding to both classes, TRKM effectively captures complex data patterns. The kernel trick is employed to project data into a high-dimensional feature space, where an optimal separating hyperplane is identified using a regularized least squares approach, enhancing both performance and computational efficiency. Extensive experiments on 36 diverse datasets from UCI and KEEL repositories demonstrate TRKM’s superior accuracy and scalability compared to baseline models. Additionally, TRKM’s application to the brain age estimation dataset underscores its efficacy in predicting brain age, a critical biomarker for early Alzheimer’s disease detection, highlighting its potential for real-world medical applications. To the best of our knowledge, TRKM is the first twin variant of the RKM framework, offering a robust and efficient solution for complex classification and regression tasks. The source code of the proposed TRKM model is available at \url{https://github.com/mtanveer1/TRKM}.
\end{abstract}

\end{frontmatter}

\section{Introduction}
Support vector machines (SVMs) \cite{cortes1995support} have emerged as a powerful tool for solving classification problems. SVMs are based on statistical learning theory and the principle of maximizing the margin to identify the optimal hyperplane that separates classes, achieved by solving a quadratic programming problem (QPP). SVM has been extensively applied to a wide range of real-world challenges, such as remote sensing \cite{pal2005support}, EEG signal classification \cite{richhariya2018eeg}, diagnosis of Alzheimer's disease \cite{quadir2024granular}, feature extraction \cite{li2008joint}, and so on. SVM offers significant advantages by employing the structural risk minimization (SRM) principle, which enhances generalization and reduces errors during the training phase. However, it faces challenges in real-world applications due to its high computational complexity. Least squares SVM (LSSVM) was proposed by \citet{suykens1999least} to address the computational burden of SVM. Unlike SVM, LSSVM uses a quadratic loss function rather than a hinge loss function, enabling the use of equality constraints in the classification problem. As a result, the solution can be obtained by solving a system of linear equations, thereby avoiding the need to solve a large QPP. LSSVM has significantly lower computation time than SVM, making it more efficient for large-scale problems. Over the past decade, significant progress has been made in enhancing the accuracy of SVM. One notable development is the generalized eigenvalue proximal SVM (GEPSVM) introduced by \citet{mangasarian2005multisurface}, and the twin SVM (TSVM), proposed by \citet{khemchandani2007twin}. GEPSVM addresses the generalized eigenvalue problem, avoiding the need to solve a QPP. However, TSVM is four times faster than standard SVM because it solves two smaller QPPs rather than a single large QPP \cite{khemchandani2007twin, tanveer2022comprehensive}, establishing it as a notable and superior alternative. In order to keep each hyperplane near the data points of one class and at least one unit away from the data points of the other class, TSVM generates a pair of non-parallel hyperplanes. Recently, several variants of TSVM have been introduced, such as the enhanced feature based granular ball TSVM (EF-GBTSVM) \cite{quadir2024enhanced}, least squares TSVM (LSTSVM) \cite{kumar2009least}, large-scale pinball TSVM \cite{tanveer2022large}, intuitionistic fuzzy weighted least squares TSVM \cite{tanveer2022intuitionistic}, multiview SVM with wave loss (Wave-MvSVM) \cite{quadir2024enhancing} and many more.

While SVM and its variants are effectively applied to regression tasks, they are referred to as support vector regression (SVR) \cite{basak2007support}. In contrast to SVM, SVR generates a tolerance margin \( \epsilon \) and seeks the best hyperplane to minimize the error within it. Thus, SVR identifies a function where the error can be up to \( \epsilon \) distance, meaning any error within \( \epsilon \) deviation from the true values is considered acceptable. SVR learns at a relatively slow pace because it requires solving a QPP. Researchers have developed many variants of SVR to boost its performance by lowering computational complexity and increasing accuracy \cite{drucker1996support}. SVR involves high computational costs, which led to the development of the efficient twin SVR (TSVR) \cite{peng2010tsvr}. TSVR utilizes \( \epsilon \)-insensitive upper and lower bounds to refine and optimize the final regression function. Its formulation is more computationally efficient, as it requires solving two smaller QPPs, leading to a substantial increase in speed compared to SVR. Further, several variants of TSVR have been proposed, including twin projection SVR \cite{peng2014twin}, \( \epsilon \)-TSVR \cite{shao2013varepsilon}, and twin support vector quantile regression (TSVQR) \cite{ye2024twin}, and so on. The previously described variations solve a pair of QPPs, which may need a significant amount of memory and time. \citet{zhao2013twin} proposed twin least squares SVR (TLSSVR) to address this high computational burden. TLSSVR replaces the inequality constraints of TSVR with equality constraints, which allows the model to be trained by solving a system of linear equations.

Linear models such as SVM and SVR, along with their extensions, inherently produce linear decision boundaries. As a result, they often struggle to achieve high classification accuracy when dealing with data that is not linearly separable. In contrast, kernel-based variants of SVM \cite{scholkopf2002learning, si2025kernel} and SVR \cite{basak2007support} have demonstrated remarkable success across diverse application areas. These methods excel at modeling complex data structures by capturing non-linear relationships, thereby enhancing performance in both classification and regression tasks \cite{shawe2004kernel, bishop2006pattern}. The core strength of kernel methods lies in their ability to implicitly project data into a high-dimensional feature space, allowing for the effective learning of non-linear patterns without explicitly computing the transformation. This capability makes kernel-based SVM and SVR highly adaptable to a wide range of data types and problem domains \cite{cristianini2000introduction}. Nonetheless, kernel methods face scalability challenges when applied to large-scale datasets, as computational costs grow significantly with dataset size. Moreover, choosing an appropriate kernel function and tuning its hyperparameters is a non-trivial process that requires careful experimentation and domain expertise. To address some of these challenges, \citet{suykens2017deep} proposed the restricted kernel machine (RKM), which blends the principles of kernel methods with neural network paradigms. This innovation enhances the applicability of kernel techniques to more intricate real-world scenarios. RKM offers a reformulation of the least square SVM (LSSVM) framework by leveraging the Legendre-Fenchel duality \cite{rockafellar1974conjugate}, and structurally resembles the energy-based formulation of restricted boltzmann machines (RBM) \cite{hinton2006fast}. RKM has been successfully applied to various tasks, including generative modeling \cite{pandey2021generative, pandey2022disentangled}, classification \cite{houthuys2021tensor, quadir2025one, quadir2025randomized}, and learning disentangled representations \cite{tonin2021unsupervised}. By using kernel mappings, RKM constructs non-linear decision surfaces in a high-dimensional space, enabling it to model complex feature interactions effectively. However, as with traditional kernel methods, RKM's computational requirements increase substantially with the size of the dataset, posing challenges for scalability in large-scale applications.

The rapid growth of data complexity and volume in modern machine learning applications demands models that balance computational efficiency, generalization performance, and robustness across diverse tasks. While kernel-based methods, such as the RKM, leverage the kernel trick to handle non-linear data structures effectively, they often face computational challenges and scalability issues when applied to large datasets. Additionally, traditional support vector machine variants like TSVM and LSTSVM have shown promise in improving efficiency and generalization through twin hyperplane approaches. Yet, they lack the flexibility to exploit the latent feature representations inherent in kernel methods fully. To address these limitations, we propose the twin restricted kernel machine (TRKM), a novel framework that integrates the strengths of RKM with the computational and generalization advantages of twin methods. TRKM introduces a twin variant of RKM, combining the expressive power of kernel-based methods with the efficiency of twin hyperplane strategies, making it suitable for both classification and regression tasks. By incorporating an energy function inspired by restricted Boltzmann machines and leveraging conjugate feature duality based on the Fenchel-Young inequality, TRKM captures complex data patterns while maintaining computational tractability. The model’s ability to project data into a high-dimensional feature space and identify optimal hyperplanes through a regularized least squares approach further enhances its adaptability and performance.

Our motivation stems from the pressing need to develop a robust, scalable, and computationally efficient learning framework that can overcome the limitations of the conventional Restricted Kernel Machine (RKM), particularly its computational bottlenecks and sensitivity to data complexity, while still preserving its powerful kernel-based representational capabilities. The proposed Twin Restricted Kernel Machine (TRKM) is specifically designed to enhance both generalization and robustness, effectively balancing structural risk minimization with computational tractability. Unlike traditional RKM formulations that solve a single optimization problem over the entire data distribution, TRKM decomposes the learning process into two smaller, coupled subproblems, enabling faster convergence and better handling of high-dimensional or large-scale datasets. Comprehensive experiments conducted on 36 benchmark datasets from the UCI and KEEL repositories, along with a real-world application in early Alzheimer’s disease detection using brain age as a biomarker, validate the scalability and adaptability of TRKM across diverse domains. To the best of our knowledge, TRKM represents the first twin formulation of the RKM framework, introducing a novel perspective that bridges the gap between energy-based kernel models and twin support vector paradigms, thereby establishing a significant step forward in the evolution of kernel-based learning for both classification and regression tasks. The main contributions of our work are as follows:
\begin{enumerate}
    \item We propose twin restricted kernel machine (TRKM), a novel model that synergizes the strengths of twin support vector machine (TSVM) with the restricted kernel machine (RKM) framework. By addressing RKM's computational and generalization challenges, TRKM provides a robust and efficient solution for both classification and regression tasks, marking the first twin variant of RKM.. 
    \item TRKM leverages an energy-based mechanism inspired by Restricted Boltzmann Machines to capture complex data patterns, enabling superior modeling of intricate relationships.
    \item By integrating kernel methods with a conjugate feature duality formulation, TRKM achieves both theoretical rigor and computational efficiency, offering improved generalization and stability.
    \item We introduce a conjugate feature duality using the Fenchel-Young inequality, reformulating classification and regression in terms of dual variables. This establishes a bound on the TRKM objective, providing a novel approach that improves computational efficiency and robustness.
    \item We conducted extensive experiments on 36 real-world datasets from the UCI and KEEL repositories, along with additional evaluations on large-scale NDC and real-world benchmark datasets. Through rigorous numerical experiments and comprehensive statistical analyses, the TRKM model consistently outperforms baseline methods, demonstrating its scalability, robustness, and strong generalization capability across both moderate- and large-scale learning scenarios.
    \item TRKM is evaluated on the brain age dataset, where it exhibited strong empirical performance in the early detection of Alzheimer’s disease using brain age as a biomarker. These findings highlight TRKM’s potential for impactful real-world applications in medical diagnostics.
\end{enumerate}

The rest of this paper is structured as follows. Section \ref{Related Works} reviews related work in the field. Sections \ref{TRKM-C} present the mathematical formulations for the proposed TRKM for classification and regression, respectively. Section \ref{Discussion of the proposed TRKM model w.r.t. the baseline models} provides the discussion of the proposed TRKM model against the baseline models. Experimental results are analyzed in Section \ref{Experiments and Results}. Finally, Section \ref{Conclusion} concludes the article and outlines potential directions for future research.

\section{Related Works}
\label{Related Works}
This section presents the notations used in this paper, along with a brief overview of TSVM, TSVR, and the proposed RKM model for classification and regression.

\subsection{Notations}
Consider a classification dataset denoted by \(\mathscr{T}\), consisting of pairs \((x_i, y_i)\) where \(x_i \in \mathbb{R}^{1 \times m}\) is a feature vector with \(m\) features and \(y_i \in \{+1, -1\}\) is the class label, with \(+1\) indicating the positive class and \(-1\) indicating the negative class. The positive class feature vectors are collected into an \(n_1 \times m\) matrix \(A\), where \(n_1\) is the number of positive samples, and the negative class feature vectors are collected into an \(n_2 \times m\) matrix \(B\), where \(n_2\) is the number of negative samples. Here, \(n = n_1 + n_2\) represents the total number of samples in the dataset. \( e_1 \in \mathbb{R}^{n_1 \times 1} \), \( e_2 \in \mathbb{R}^{n_2 \times 1} \) and $e \in \mathbb{R}^{n \times 1}$ are column vectors of ones. We use the Gaussian kernel, which is defined as \( k(x_i, x_j) = e^{-\frac{1}{2\sigma^2} \| x_i - x_j \|^2} \), which allows us to construct a kernel matrix \( \mathcal{K} \) for the input data:
\begin{align}
    \mathcal{K} =  
\begin{bmatrix}  
k(x_1, x_1) & \cdots & k(x_1, x_n) \\  
\vdots & \ddots & \vdots \\  
k(x_n, x_1) & \cdots & k(x_n, x_n)  
\end{bmatrix},
\end{align}
where $\sigma$ is the kernel hyperparameter.

Consider a regression dataset denoted by \(\mathscr{M} \), consisting of pairs \((x_i, t_i)\) where \(x_i \in \mathbb{R}^{1 \times m}\) is a feature vector with \(m\) features and \(t_i \in \mathbb{R}\) is the corresponding continuous output value. The feature vectors \(x_i\) are organized into an \(n \times m\) matrix \(X\), where \(n\) is the total number of samples, and the output values \(t_i\) are organized into an \(n \times 1\) vector \(Y\). Here, \(n\) represents the total number of samples in the regression dataset. For clarity, Table \ref{Descriptions of key notations} provides an overview of the key symbols used throughout this paper.

\begin{table*}[ht!]
\centering
    \caption{Descriptions of key notations.}
    \label{Descriptions of key notations}
    \resizebox{0.9\linewidth}{!}{
\begin{tabular}{ll}
\hline
Notations & Descriptions \\ \hline
 $n$  & the number of samples. \\
 $x_i$  & An $i^{th}$ training sample. \\
 $y_i$  & The class label for the $i^{th}$ input feature vector (e.g., $+1$ or $-1$). \\
 $t_i$ & The target output (dependent variable) for $x_i$. \\
 $w_i$  & The weight vector. \\ 
 $bi$ & The bias term. \\
 $\phi$ & The high-dimensional mapping. \\
 $k(x_i, x_j)$ & The kernel function. \\
 $\mathcal{K}$ & Kernel matrix. \\
 $e_i$ & The vector of ones of appropriate dimension. \\
 $A$  & Matrix of training samples belonging to class $+1$. \\
 $B$ & Matrix of training samples belonging to class $-1$. \\
 $n_1$ & Number of samples in class $+1$. \\
 $n_2$ & Number of samples in class $-1$. \\ 
 $m$ & Dimensionality of the input feature space. \\
 $\|\cdot\|$ & $L_2$ norm (Euclidean norm) of a vector. \\
 $I$ & Identity matrix.\\
 $h_i$ & Hidden feature vector (hidden unit). \\
 $Tr(\cdot)$ &  Trace of a matrix. \\
 $J_i$ & Objective function (primal form). \\
$d_i$, $\gamma_i$, and $\eta_i$ & Regularization parameter, i.e., controls the trade-off between fitting the data and model complexity. \\
$\xi_i$ & Slack variable vector. \\ \hline
 \end{tabular}}
\end{table*}

\subsection{Twin Support Vector Machine (TSVM)}
In TSVM \cite{khemchandani2007twin}, two non-parallel hyperplanes are generated. Each hyperplane passes through the samples of its respective class and aims to maximize the margin between the hyperplanes and the samples of the opposing class. The optimization problem for TSVM can be formulated as follows:
\begin{align}
\label{eq:1}
      & \underset{w_1,b_1}{min} \hspace{0.2cm} \frac{1}{2} \|\mathcal{K}(A, C^T)w_1 + e_1b_1\|^2 + d_1e_2^T\xi_2 \nonumber \\
     & s.t. \hspace{0.2cm} -(\mathcal{K}(B, C^T)w_1 + e_2b_1) + \xi_2 \geq e_2, \nonumber \\
     & \hspace{0.8cm} \xi_2 \geq 0,
\end{align}
and
\begin{align}
\label{eq:2}
    & \underset{w_2,b_2}{min} \hspace{0.2cm} \frac{1}{2} \|\mathcal{K}(B, C^T)w_2 + e_2b_2\|^2 + d_2e_1^T\xi_1 \nonumber \\
     & s.t. \hspace{0.2cm} (\mathcal{K}(A, C^T)w_2 + e_1b_2) + \xi_1 \geq e_1, \nonumber \\
     & \hspace{0.8cm} \xi_1 \geq 0,
\end{align}
where \( d_1 \) and \( d_2 \) (\( > 0 \)) are penalty parameters, \( e_1 \in \mathbb{R}^{n_1 \times 1} \) and \( e_2 \in \mathbb{R}^{n_2 \times 1} \) are column vectors of ones with appropriate dimensions, $\mathcal{K}$ is kernel function, $C = [A; B]$ and \( \xi_1 \) and \( \xi_2 \) are slack vectors, respectively.

\subsection{Twin Support Vector Regression (TSVR)}
TSVR extends the concept of TSVM to regression problems. In TSVR, two separate regression functions are learned, each focusing on approximating different parts of the data distribution. By solving two QPPs simultaneously, TSVR aims to find two distinct regression functions that minimize the prediction error while maintaining a balance between the models, which are defined as:
\begin{align}
    f_1(x) = w_1^T\phi(x) + b_1  ~~~ \text{and} ~~~ f_2(x) = w_2^T\phi(x) + b_2.
\end{align}
The optimization problem of TSVR is given as follows:
\begin{align}
\label{eq:5}
      & \underset{w_1,b_1}{min} \hspace{0.2cm} \frac{1}{2} \|Y - e\epsilon_1 - (\mathcal{K}(X, X^T)w_1 + e_1b_1)\|^2 + d_1e^T\xi_2 \nonumber \\
     & s.t. \hspace{0.2cm} Y-(\mathcal{K}(X, X^T)w_1 + eb_1) + \xi_2 \geq e \epsilon_1, \nonumber \\
     & \hspace{0.8cm} \xi_2 \geq 0,
\end{align}
and
\begin{align}
\label{eq:6}
    & \underset{w_2,b_2}{min} \hspace{0.2cm} \frac{1}{2} \|Y + e\epsilon_2 - (\mathcal{K}(X, X^T)w_2 + eb_2)\|^2 + d_2e^T\xi_1 \nonumber \\
     & s.t. \hspace{0.2cm} (\mathcal{K}(X, X^T)w_2 + e_1b_2) - Y + \xi_1 \geq e\epsilon_2, \nonumber \\
     & \hspace{0.8cm} \xi_1 \geq 0,
\end{align}
where \( d_1 \) and \( d_2 \) (\( > 0 \)) are regularization parameters, $\xi_1$ and $\xi_2$ are slack vectors, respectively. Therefore, the regression function of a nonlinear TSVR can be expressed as:
\begin{align}
\label{eq:7}
    f(x) &= \frac{1}{2}(f_1(x) + f_2(x)) \nonumber \\
   & = \frac{1}{2}((w_1+w_2)^T\mathcal{K}(x, X^T)) +\frac{1}{2}(b_1 + b_2).
\end{align}
\subsection{Restricted Kernel Machine (RKM)}
Here, we present an overview of the RKM model, as described by \citet{suykens2017deep}, which is closely related to the well-known LSSVM model \cite{suykens1999least}. RKM utilizes the kernel trick to transform the data into a high-dimensional feature space, enabling the construction of a linear separating hyperplane. The optimization problem for RKM is given as follows:
\begin{align}
\label{eq:8}
      J = & \frac{\gamma}{2} Tr(w^Tw) + \sum_{i=1}^n (1-(\phi(x_i)^Tw+b)y_i)h_i - \frac{\eta}{2} \sum_{i=1}^nh_i^2,
\end{align}
where \(\gamma\) and \(\eta\) are the regularization parameters, \(b\) is the bias term, and \(h\) represents the hidden features. The solution to equation \eqref{eq:8} is obtained by taking the partial derivatives of \(J\) with respect to (w.r.t.) \(w\), \(b\), and \(h_i\), and then setting these derivatives to zero. For a detailed derivation, refer to \citet{suykens2017deep}. The solution to the optimization problem is given by:
\begin{align}
    \begin{bmatrix}
e & \frac{1}{\gamma} \mathcal{K} + \eta I \\
0  & e^T
\end{bmatrix}
\begin{bmatrix}
b \\
h
\end{bmatrix}
= 
\begin{bmatrix}
y \\
0
\end{bmatrix},
\end{align}
where $\mathcal{K}$ is the kernel matrix, \(e \in \mathbb{R}^{n \times 1}\) denotes a column vector with all its entries equal to one, $I$ is the identity matrix, and $h \in \mathbb{R}^{n \times 1}$.

\section{Proposed Twin Restricted Kernel Machine (TRKM)}
\label{TRKM-C}
In this section, we proposed the twin restricted kernel machine (TRKM) model to address both classification and regression tasks. Although these tasks have inherently different architectures, objectives, and evaluation criteria, our approach is built upon a unified mathematical framework that maintains structural consistency across both problem domains. Specifically, we introduce two distinct architectures tailored to classification and regression, respectively; however, both architectures are formulated within the same conceptual foundation inspired by the RKM.
\\
The core idea behind TRKM is to extend the RKM framework in a way that accommodates the twin hyperplane formulation, enabling the model to handle classification and regression tasks in a parallel and consistent manner. For classification, the TRKM constructs two non-parallel decision surfaces to better separate data classes, while for regression, it models two parallel functions to bracket the target outputs and reduce prediction error. Despite these architectural differences, the underlying optimization principles, kernel mappings, and dual formulations remain largely consistent across both tasks.
\\
This shared mathematical structure not only simplifies the theoretical formulation of TRKM but also ensures computational efficiency and ease of implementation. By leveraging this unified approach, TRKM demonstrates flexibility and robustness in adaptation classification and regression.

\subsection{Proposed Twin Restricted Kernel Machines for Classification (TRKM-C)}
This section provides a comprehensive explanation of the proposed twin restricted kernel machine for classification (TRKM-C), a novel framework designed to advance kernel-based classification by integrating the computational efficiency of twin methods with the expressive power of RKM. TRKM-C introduces a unique approach by incorporating both visible and hidden variables within a non-probabilistic energy function, inspired by restricted Boltzmann machines (RBM) \cite{hinton2006fast}. This energy-based formulation establishes a bridge between kernel methods and RBM, enabling TRKM-C to capture complex, non-linear data patterns while addressing the computational and scalability challenges inherent in traditional RKM and other kernel-based methods. Unlike conventional kernel approaches, TRKM-C leverages a twin hyperplane strategy, akin to TSVM and LSTSVM, to split the classification task into two smaller optimization problems, each corresponding to one of the $+1$ or negative $-1$ classes, represented by visible units $v_1$ ($v_2$) and hidden units $h_1$ ($h_2$). 
\\
The TRKM-C framework employs a scalar-valued energy function that governs both the training and prediction phases, similar to the energy function in RBM. This energy function facilitates the modeling of intricate relationships between visible and hidden variables, allowing TRKM-C to represent complex data distributions effectively. To achieve robust classification, TRKM-C utilizes the kernel trick, mapping input samples from the original space to a high-dimensional feature space via the function ($\phi: x_i \rightarrow \phi(x_i)$). In this feature space, TRKM-C identifies two optimal hyperplanes, one for each class, using a regularized least squares approach that ensures efficient separation of training instances while maintaining generalization performance. The optimization process is streamlined by solving a linear system or performing matrix decomposition, which significantly reduces the computational complexity compared to traditional RKM, which often requires solving computationally intensive quadratic programming problems. A key innovation in TRKM-C is the incorporation of conjugate feature duality based on the Fenchel-Young inequality, which reformulates the classification problem in terms of conjugate dual variables. This duality provides a bound on the TRKM-C objective function, enhancing the model’s robustness and computational efficiency by simplifying the optimization landscape.

The formulation of TRKM-C for the first hyperplane is given as follows:
\begin{align}
\label{eq:9}
     \underset{w_1, \xi_1}{\min}~J_1(w_1, \xi_1) = & \underset{w_1, \xi_1}{\min}~\frac{\gamma_1}{2} w_1^Tw_1  + e_2^T(\phi(B)w_1 + e_2b_1) +\frac{1}{2\eta_1}\xi_1^T\xi_1 \nonumber \\
   & \hspace{0.07cm} \textbf{s.t.}  \hspace{0.2cm} \xi_1 = e_1 - \phi(A)w_1 - e_1 b_1, 
\end{align}
where $w_1 \in \mathbb{R}^{n_2 \times 1}$ is the interconnection vector, $\xi_1 \in \mathbb{R}^{n_1 \times 1}$ is the error vector, \( b_1 \) denotes the bias term, the vectors $e_1$ and $e_2$ are ones of the suitable dimensions, and \( \gamma_1 \) and \( \eta_1 \) are the tunable parameters, respectively. The objective function \ref{eq:9} consists of three key components, each playing a crucial role in model optimization. The first term controls model complexity while determining the positive hyperplane, ensuring a balance between generalization and overfitting. The second term $ e_2^T (\phi(B) w_1 + e_2 b_1) $ represents the margin term for the second class (class $-1$) in the feature space induced by the kernel mapping $\phi$. It measures how well the hyperplane defined by $w_1$ and bias $b_1$ separates the samples of class $-1$ from the decision boundary for class $+1$. Specifically, it encourages the samples of class $-1$ to lie on the correct side of the non-parallel hyperplane associated with class $+1$, maximizing the margin between classes. This term is critical for the twin model structure, as it ensures that the hyperplane for class $+1$ (defined by $\phi(A) w_1 + e_1 b_1$) is optimized to push class $-1$ samples away, while the constraint $\xi_1 = e_1 - \phi(A) w_1 - e_1 b_1$ ensures that class $+1$ samples are correctly classified up to slack variables $\xi_1$. The use of $e_2^T$ (a vector of ones for class $-1$ samples) aggregates the margin contributions across all class $-1$ samples, aligning with the goal of non-parallel hyperplane-based classification. Lastly, the third term penalizes the sum of error variables, reducing the risk of excessive overfitting to the positive training points.

The connection between LSSVM and RBM in TRKM-C is formed by applying the Legendre-Fenchel conjugate for quadratic functions \cite{rockafellar1974conjugate}. The TRKM-C establishes a bound for \( J_1 \) by incorporating the hidden layer representations \( h_1 \) as follows:
\begin{align}
\label{eq:10}
    \frac{1}{2\eta_1}\xi_1^T\xi_1 \geq \xi_1^Th_1 - \frac{\eta_1}{2}h_1^Th_1, \hspace{0.2cm} \forall \hspace{0.1cm} \xi_1, h_1.
\end{align}
Combining \eqref{eq:9} and \eqref{eq:10} leads to
\begin{align}
\label{eq:11}
    J_1 \leq ~& \xi_1^T h_1 - \frac{\eta_1}{2} h_1^Th_1 + \frac{\gamma_1}{2} w_1^Tw_1 + e_2^T(\phi(B)w_1 + e_2b_1) \nonumber \\ 
   &   \textbf{s.t.} \hspace{0.2cm} \xi_1 = e_1 - \phi(A)w_1 - e_1 b_1.
\end{align}
The tight bound of $J_1$ can be obtained by incorporating the constraints as follows:
\begin{align}
\label{eq:12}
    J_1 \leq &  (e_1 - \phi(A)w_1 - e_1 b_1)^T h_1 - \frac{\eta_1}{2} h_1^Th_1 + \frac{\gamma_1}{2} w_1^Tw_1 + e_2^T(\phi(B)w_1 + e_2b_1) \nonumber \\
    & =  \hat{J}_1.
\end{align}

We introduce hidden features ($h_1$) through conjugate feature duality, leveraging the mathematical concept of Fenchel duality to conjugate these features to the error variables $\xi_1$, defined as $\xi_1 = e_1 - \phi(A)w_1 - e_1 b_1$, where $\xi_1$ represents the error term, $\phi(A)$ is the feature mapping of the input data. The resulting inner pairing term $\xi_1^T h_1 = (e_1 - \phi(A)w_1 - e_1 b_1)^T h_1$ mirrors the interaction term in the energy function of an RBM, enabling TRKM to capture complex data representations. This formulation establishes a robust framework that bridges kernel methods with energy-based models, enhancing the model's ability to learn intricate patterns in high-dimensional feature spaces for classification tasks.

Now, examine the stationary points of $\hat{J}_1$ by taking the gradients w.r.t. $w_1$, $h_1$, and $b_1$ as follows:
\begin{align}
    & \frac{\partial \hat{J}_1}{\partial w_1} = 0 \implies w_1 = \frac{1}{\gamma_1} (\phi(A)^Th_1 - \phi(B)^Te_2), \label{eq:13} \\
    & \frac{\partial \hat{J}_1}{\partial h_1} = 0 \implies \eta_1 h_1 = e_1 - \phi(A)w_1 - e_1 b_1, \label{eq:14}  \\
    & \frac{\partial \hat{J}_1}{\partial b_1} = 0 \implies h_1^Te_1 = e_2^te_2. \label{eq:15}
\end{align}
Substituting the weight vector \( w_1 \) from Eq \eqref{eq:13} into \eqref{eq:14}, we obtain:
\begin{align}
\label{eq:16}
    \frac{1}{\gamma_1} \phi(A)\phi(A)^Th_1 - \frac{1}{\gamma_1} \phi(A)\phi(B)^Te_2 + \eta h_1 + b_1e_1 = e_1.
\end{align}
By calculating the stationary points of the objective function, we obtain the following system of linear equations:
\begin{align}
\label{eq:17}
\left[\begin{array}{c|c } 
	\frac{1}{\gamma_1} \mathcal{K}(A, A^T) + \eta_1 I & e_1  \\  
	\hline 
	 e_1^T    & 0  
\end{array}\right]
    \begin{bmatrix}
    h_1 \\ b_1 
    \end{bmatrix}  =  \begin{bmatrix} e_1 + \frac{1}{\gamma_1} \mathcal{K}(A, B^T)e_2 \\   e_2^Te_2   \end{bmatrix},
\end{align}
where $I$ represents a matrix of ones of appropriate dimension, and $\mathcal{K}$ represents the kernel function.

The optimization problem of TRKM-C for the second hyperplane is given as follows:
\begin{align}
\label{eq:18}
     \underset{w_2, \xi_2}{\min}~J_2(w_2, \xi_2) = & \underset{w_2, \xi_2}{\min}~\frac{\gamma_2}{2} w_2^Tw_2  - e_1^T(\phi(A)w_2 + e_1b_2) +\frac{1}{2\eta_2}\xi_2^T\xi_2 \nonumber \\
   & \hspace{0.07cm} \textbf{s.t.}  \hspace{0.2cm} \xi_2 = -e_2 - \phi(B)w_2 - e_2 b_2.
\end{align}
The tight bound of $J_2$ can be obtained by incorporating the constraints as follows:
\begin{align}
\label{eq:19}
    J_2 \leq &  (-e_2 - \phi(B)w_2 - e_2 b_2)^T h_2 - \frac{\eta_2}{2} h_2^Th_2 + \frac{\gamma_2}{2} w_2^Tw_2 - e_1^T(\phi(A)w_2 + e_1b_2) \nonumber \\
    & =  \hat{J}_2.
\end{align}
Analogously, $w_2$ corresponding to the $-1$ class can be calculated by the subsequent equation:
\begin{align}
\label{eq:20}
    w_2 = \frac{1}{\gamma_2} (\phi(B)^Th_2 + \phi(A)^Te_1).
\end{align}
By determining the stationary points of the objective for the \(-1\) class, we obtain the following linear problem:
\begin{align}
\label{eq:21}
\left[\begin{array}{c|c } 
	\frac{1}{\gamma_2} \mathcal{K}(B, B^T) + \eta_2 I & e_2  \\  
	\hline 
	 e_2^T    & 0  
\end{array}\right]
    \begin{bmatrix}
    h_2 \\ b_2
    \end{bmatrix}  = - \begin{bmatrix} e_2 + \frac{1}{\gamma_2} \mathcal{K}(B, A^T)e_1 \\   e_1^Te_1   \end{bmatrix}.
\end{align}

Once the optimal values of $h_1$ ($b_1$) and $h_2$ ($b_2$) are calculated for the $+1$ and $-1$ class, respectively. To predict the label of a new sample \( x \), the following decision function can be used as follows:
\begin{align}
\label{eq:22}
    \text{class}(x) =  sign\left(g_1(x) + g_2(x)\right),
\end{align}
where
\begin{align}
    g_1(x) = \frac{1}{\gamma_1}[\mathcal{K}(x,A^T)h_1 - \mathcal{K}(x,B^T)e_2] + b_1, \label{eq:23}
\end{align}
and
\begin{align}
    g_2(x) = \frac{1}{\gamma_2}[\mathcal{K}(x,B^T)h_2 + \mathcal{K}(x,A^T)e_1] + b_2. \label{eq:24} 
\end{align}
Algorithm \ref{TRKM-C training and prediction} provides a concise description of the proposed TRKM-C algorithm.

\subsection{Proposed Twin Restricted Kernel Machines for Regression (TRKM-R)}
This section provides a detailed formulation of the proposed twin restricted kernel machines for regression (TRKM-R). TRKM-R can be related to the energy form expression of an RBM by interpreting it in terms of the hidden and visible units. This connection enables us to interpret TRKM-R using a framework similar to RBMs, where the energy function captures the interactions between hidden and visible units. Given these parallels with RBMs and the absence of hidden-to-hidden connections, we refer to this specific interpretation of the model as a TRKM representation for regression. The first optimization problem of TRKM-R can be obtained as:
\begin{align}
\label{eq:25}
     \underset{w_1, \xi_1}{\min}~J_1(w_1, \xi_1) = & \underset{w_1, \xi_1}{\min}~\frac{\gamma_1}{2} w_1^Tw_1  -  e^T(\phi(X)w_1 + eb_1) +\frac{1}{2\eta_1}\xi_1^T\xi_1 \nonumber \\
   & \hspace{0.07cm} \textbf{s.t.}  \hspace{0.2cm} Y = \phi(X)w_1 + eb_1 - \xi_1. 
\end{align}

We now analyze the optimization problem in Equation \ref{eq:25}. The first term in the objective function, \(\frac{1}{2} \text{Tr}(w_1^T w_1)\), regulates the complexity of the function \( f_1(x) \), ensuring a balance between model expressiveness and overfitting. The second term $ - e^T (\phi(X) w_1 + e b_1) $ represents the negative sum of the predicted outputs for the input data $X$ in the feature space. This term encourages the model’s predictions ($\phi(X) w_1 + e b_1$) to align closely with the true target values $Y$, as the constraint $Y = \phi(X) w_1 + e b_1 - \xi_1$ defines the residual errors $\xi_1$. The negative sign ensures that the objective function minimizes the deviation of predictions from the true values, effectively acting as a fidelity term to the target data. This term is essential for regression, as it directly ties the model’s predictions to the observed data, balancing the trade-off between model complexity (first term: $\frac{\gamma_1}{2} \text{Tr}(w_1^T w_1)$) and prediction accuracy (third term: $\frac{1}{2 \eta_1} \xi_1^T \xi_1$). Unlike TRKM-C, which focuses on margin maximization for classification, TRKM-R’s second term prioritizes accurate prediction of continuous outputs, making it suitable for regression tasks. The constraints enforce that the estimated values obtained by \( f_1(x) \) remain at a unit distance from the corresponding response values of the training points. If this condition is not satisfied, a slack vector \( \xi_1 \) is introduced to quantify the deviation. Lastly, the third term in the objective function penalizes the sum of error variables, mitigating the risk of excessive overfitting to the training data.

We derived the bound for \( J \) by applying the same property used in the classification problem, resulting in:
\begin{align}
    &J_1 \leq \xi_1^T h_1 - \frac{\eta_1}{2} h_1^Th_1 + \frac{\gamma_1}{2} w_1^Tw_1 - e^T(\phi(X)w_1 + eb_1) \nonumber \\ 
   &  \textbf{s.t.} \hspace{0.2cm}  Y = \phi(X)w_1 + eb_1 - \xi_1.
\end{align}
The bound can be obtained by substituting the constraints:
\begin{align}
    J_1 \leq &  (\phi(X)w_1 + eb_1 - Y)^T h_1 - \frac{\eta_1}{2} h_1^Th_1 + \frac{\gamma_1}{2} w_1^Tw_1 - e^T(\phi(X)w_1 + eb_1) \nonumber \\
   & =  \hat{J_1}.
\end{align}
The stationary points of $\hat{J_1}$ are given by 
\begin{align}
    & \frac{\partial \hat{J}_1}{\partial w_1} = 0 \implies w_1 = \frac{1}{\gamma_1} (\phi(X)^Te - \phi(X)^th_1), \label{eq:28} \\
    & \frac{\partial \hat{J}_1}{\partial h_1} = 0 \implies \eta_1 h_1 = \phi(X)w_1 + e b_1 - Y, \label{eq:29}  \\
    & \frac{\partial \hat{J}_1}{\partial b_1} = 0 \implies h_1^Te = e^Te. \label{eq:30}
\end{align}
On employing $w_1$ in \eqref{eq:29}, we get
\begin{align}
\label{eq:31}
    \frac{1}{\gamma_1} \phi(X)\phi(X)^Te - \frac{1}{\gamma_1} \phi(X)\phi(X)^Th_1 - \eta h_1 + eb_1 = Y.
\end{align}
Using Eqs \eqref{eq:31} and \eqref{eq:30}, we can find the solution by solving the following system of linear equations:
\begin{align}
\label{eq:32}
\left[\begin{array}{c|c } 
	\frac{1}{\gamma_1} \mathcal{K}(X, X^T) + \eta_1 I & -e  \\  
	\hline 
	 e^T    & 0  
\end{array}\right]
    \begin{bmatrix}
    h_1 \\ b_1 
    \end{bmatrix}  =  \begin{bmatrix} -Y + \frac{1}{\gamma_1} \mathcal{K}(X, X^T)e \\   e^Te   \end{bmatrix}.
\end{align}
The second optimization problem of TRKM-R can be obtained as follows:
\begin{align}
\label{eq:33}
     \underset{w_2, \xi_2}{\min}~J_2(w_2, \xi_2) = & \underset{w_2, \xi_2}{\min}~\frac{\gamma_2}{2} w_2^Tw_2  +  e^T(\phi(X)w_2 + eb_2) +\frac{1}{2\eta_2}\xi_2^T\xi_2 \nonumber \\
   & \hspace{0.07cm} \textbf{s.t.}  \hspace{0.2cm} Y = \phi(X)w_1 + eb_2 + \xi_2. 
\end{align}
By applying the constraints, we derive the following accurate bound for \( J_2 \):
\begin{align}
\label{eq:34}
    J_2 \leq &  (Y - \phi(X)w_2 - eb_2 )^T h_2 - \frac{\eta_2}{2} h_2^Th_2 + \frac{\gamma_2}{2} w_2^Tw_2 + e^T(\phi(X)w_2 + eb_2) \nonumber \\
    & =  \hat{J}.
\end{align}
By taking the gradient of \eqref{eq:34}, we determine the weight vector associated with the second optimization problem:
\begin{align}
    w_2 = \frac{1}{\gamma_2} (\phi(X)^Th_2 - \phi(X)^Te).
\end{align}
Similarly, we can derive the following system of linear equations:
\begin{align}
\label{eq:36}
\left[\begin{array}{c|c } 
	\frac{1}{\gamma_2} \mathcal{K}(X, X) + \eta_2 I & e  \\  
	\hline 
	 e^T    & 0  
\end{array}\right]
    \begin{bmatrix}
    h_2 \\ b_2 
    \end{bmatrix}  =  \begin{bmatrix} Y + \frac{1}{\gamma_2} \mathcal{K}(X, X^T)e \\   e^Te   \end{bmatrix}.
\end{align}
Once the optimal values of $h_1$ ($b_1$) and $h_2$ ($b_2$) are calculated. Then, we construct the estimated regressor as follows:
\begin{align}
\label{eq:37}
    g(x) =  \left(g_1(x) + g_2(x)\right)/2,
\end{align}
where
\begin{align}
    g_1(x) = \frac{1}{\gamma_1} \mathcal{K}(x, X^T)(e - h_1) + b_1, \label{eq:38}
\end{align}
and
\begin{align}
    g_2(x) = \frac{1}{\gamma_2} \mathcal{K}(x, X^T)(h_2 - e) + b_2.  \label{eq:39}
\end{align}
Algorithm \ref{TRKM-R training and prediction} provides a concise description of the proposed TRKM-R algorithm.

\begin{algorithm}
\caption{Training and prediction of the proposed TRKM-C.}
\label{TRKM-C training and prediction}
\textbf{Require:} Let $\{x_i\}_{i=1}^n$ be the input training dataset, $A$ and $B$ represent the matrix of $+1$ and $-1$ classes and the trade-off parameters $\gamma_1$, $\gamma_2$, $\eta_1$ and $\eta_2$, respectively.
\begin{algorithmic}[1]
\STATE Find the kernel matrix $\mathcal{K}(A, A^T)$, $\mathcal{K}(A, B^T)$, $\mathcal{K}(B, B^T)$, and $\mathcal{K}(B, A^T)$.\\
\STATE Calculate $h_1$, $b_1$ and $h_2$, $b_2$ corresponding to $+1$ and $-1$ class using Eqs \eqref{eq:17} and \eqref{eq:21}.\\
\STATE Find the decision function with dual representation using Eqs \eqref{eq:23} and \eqref{eq:24}.
\STATE The classification of a test sample \(x\) into class \(+1\) or \(-1\) is determined using Eq. \eqref{eq:22}.
\end{algorithmic}
\end{algorithm}

\begin{algorithm}
\caption{Training and prediction of the proposed TRKM-R.}
\label{TRKM-R training and prediction}
\textbf{Require:} Let $X$ be the input training dataset, $Y$ represent the vector that contains the input data's target values, and the trade-off parameters $\gamma_1$, $\gamma_2$, $\eta_1$, and $\eta_2$, respectively.
\begin{algorithmic}[1]
\STATE Find the kernel matrix $\mathcal{K}(X, X^T)$.\\
\STATE Calculate $h_1$, $b_1$ and $h_2$, $b_2$ using Eqs \eqref{eq:32} and \eqref{eq:36}.\\
\STATE Find the decision function with dual representation using Eqs \eqref{eq:38} and \eqref{eq:39}.
\STATE The predicted value of the new samples $x$  is determined using Eq \eqref{eq:37}.
\end{algorithmic}
\end{algorithm}

\subsection{Computational Complexity}
\label{Time Complexity}
Let \(\mathscr{T}\) represent the training dataset consisting of \(n\) samples, where \(n_1\) samples belong to class \(+1\) and \(n_2\) samples belong to class \(-1\), with each sample having \(m\) features. Unlike TSVM \cite{khemchandani2007twin}, which utilizes inequality constraints, the proposed TRKM model is formulated with equality constraints. In TRKM, the primary factors influencing computational complexity are the kernel function evaluations and matrix inversion. The kernel functions include \(\mathcal{K}(A, A^T)\), \(\mathcal{K}(B, B^T)\), and \(\mathcal{K}(A, B^T)\) (or equivalently \(\mathcal{K}(B, A^T)\)). The computational orders of these kernel evaluations are \(\mathcal{O}(n_1^2m)\), \(\mathcal{O}(n_2^2m)\), and \(\mathcal{O}(n_1n_2m)\), respectively. Consequently, the overall complexity for kernel function computations sums to \(\mathcal{O}(n^2m)\). Additionally, the resulting matrix in TRKM has a dimensionality of \((n+1) \times (n+1)\), requiring an inversion operation with a complexity of \(\mathcal{O}((n+1)^3)\). Therefore, the overall computational complexity of TRKM is \(\mathcal{O}(n^2m + (n+1)^3)\). For $n \gg m$ it can be approximated as $\mathcal{O}((n+1)^3)$.

\section{Discussion of the proposed TRKM model w.r.t. the baseline models}
\label{Discussion of the proposed TRKM model w.r.t. the baseline models}
In this section, we elucidate the comparison of the proposed TKRM model and the existing models.
\begin{enumerate}
    \item Difference between SVM and the proposed TRKM model:
        \begin{enumerate}
            \item SVM relies on solving a quadratic programming problem (QPP), which demands significant computational resources, leading to slow training times and infeasibility for large datasets. TRKM mitigates this by using a regularized least squares approach and conjugate feature duality, enabling faster and more efficient optimization.
            \item SVM struggles with memory constraints on large datasets due to the need to store and process large kernel matrices. TRKM’s twin hyperplane framework and linear system optimization reduce memory overhead, making it scalable to high-volume data.
            \item SVM’s single hyperplane approach can struggle to capture intricate non-linear patterns in complex datasets. TRKM’s integration of an RBM-inspired energy function and kernel-based feature mapping enhances its ability to model complex data relationships.
        \end{enumerate}
    \item Difference between TSVM and the proposed TRKM model:
        \begin{enumerate}
            \item TSVM requires computing matrix inversions for two smaller QPPs, leading to high memory demands that cause failures on large datasets. TRKM’s conjugate feature duality and optimized formulation reduce memory usage, ensuring robust performance across dataset sizes.
            \item TSVM’s twin hyperplane approach, while efficient, lacks the ability to model latent feature interactions as effectively as TRKM, which incorporates hidden and visible variables inspired by RBM to capture complex data structures.
            \item TSVM’s performance is highly dependent on careful parameter selection, which can be challenging. TRKM’s energy-based framework and regularized least squares approach provide greater robustness to parameter variations.
        \end{enumerate}
    \item Difference between Pin-GTSVM and the proposed TRKM model:
        \begin{enumerate}
            \item Despite using generalized pinball loss to improve robustness, Pin-GTSVM still encounters memory issues on large-scale datasets due to its reliance on kernel matrix operations. TRKM’s efficient linear system solving and twin framework enable it to handle larger datasets without memory failures.
            \item Pin-GTSVM’s loss function, while robust to noise, does not incorporate latent feature interactions as effectively as TRKM’s RBM-inspired energy function, which enhances adaptability to diverse data distributions.
            \item Pin-GTSVM’s generalization performance is constrained compared to TRKM, which leverages conjugate feature duality to achieve better separation in high-dimensional feature spaces.
        \end{enumerate}
    \item Difference between RKM and the proposed TRKM model:
        \begin{enumerate}
            \item RKM’s optimization process involves computationally intensive operations, particularly for large datasets, due to its reliance on full kernel matrix computations. TRKM addresses this by adopting a twin hyperplane approach and conjugate duality, reducing computational overhead.
            \item RKM struggles to scale to very large datasets due to memory-intensive kernel operations. TRKM’s efficient formulation ensures scalability by optimizing memory usage and computational efficiency.
            \item While RKM uses kernel methods effectively, it lacks the twin structure and energy-based latent feature integration of TRKM, which provides superior representation learning for complex classification tasks.
        \end{enumerate}
    \item Difference between GBTSVM and the proposed TRKM model:
        \begin{enumerate}
            \item GBTSVM relies on granular ball generation to reduce sample size, which may oversimplify complex data distributions. TRKM’s RBM-inspired energy function and hidden feature integration enable richer representation learning, improving performance on intricate datasets.
            \item The effectiveness of GBTSVM is sensitive to the quality and number of granular balls, which can vary across datasets. TRKM’s conjugate feature duality and kernel-based approach offer a more consistent and robust framework for diverse data types.
        \end{enumerate}
\end{enumerate}

\section{Experiments and Results}
\label{Experiments and Results}
In this section, we conduct an extensive evaluation of the proposed TRKM model by performing experiments on UCI and KEEL datasets and comparing its performance with that of leading state-of-the-art models. Moreover, we use the Brain Age prediction dataset to evaluate the proposed models.

\subsection{Experimental Setup}
The experiments are conducted in Python 3.11 on Windows 11 running on a PC with a system configuration of Intel® Xeon® Gold 6226R CPU and 128 GB of RAM. The dataset is divided randomly in a $70:30$ proportion, where $70\%$ is allocated for training the model, and the remaining $30\%$ is used for testing. We use five-fold cross-validation combined with a grid search method to fine-tune the model's hyperparameters, selecting the following ranges: \( \eta_i = \gamma_i = \{10^{-5}, 10^{-4}, \ldots, 10^5\} \) for \( i = 1, 2 \). We employ the Gaussian kernel, defined as \( k(x_i, x_j) = e^{-\frac{1}{2\sigma^2} \| x_i - x_j \|^2} \). The range \(\{2^{-5}, 2^{-4}, \ldots, 2^5\}\) is used to choose the Gaussian kernel parameter \(\sigma\). For TRKM-C and TRKM-R, we adopt equal penalty parameters, $i.e.$ $\eta_1 = \eta_2$ and $\gamma_1 = \gamma_2$, respectively. To assess the performance of the proposed TRKM-R model, we employ four metrics, including mean absolute error (MAE), root mean squared error (RMSE), negative error (Neg Error), and positive error (Pos Error). The following are the specific definitions of these metrics:
\begin{align}
   & \text{RMSE} = \sqrt{\frac{1}{n} \sum_{i=1}^n (f(x_i) - y_i)^2}, \\
    & \text{MAE} = \frac{1}{n} \sum_{i=1}^n \lvert (f(x_i) - y_i)\rvert, 
\end{align}
\begin{align}
    & \text{Pos Error} = \frac{1}{n}\sum_{i=1, f(x_i)\leq y_i }^n  \lvert (f(x_i) - y_i)\rvert, \\
    & \text{Neg Error} = \frac{1}{n}\sum_{i=1, f(x_i) > y_i }^n  \lvert (f(x_i) - y_i)\rvert.
\end{align}
 
\subsection{Experimental Results and Statistical Analysis on UCI and KEEL Datasets for Classification}

\begin{table*}[htp]
\centering
    \caption{Classification ACC and best hyperparameter of the proposed TRKM-C and the baseline models across the real-world datasets i.e., UCI and KEEL.}
    \label{Average ACC and average rank for UCI and KEEL datasets}
    \resizebox{0.9\linewidth}{!}{
\begin{tabular}{lcccccc}
\hline
Model $\rightarrow$ & SVM \cite{cortes1995support} & TSVM \cite{khemchandani2007twin} & Pin-GTSVM \cite{tanveer2019general} & RKM \cite{suykens2017deep} & GBTSVM \cite{GBquadir2024granular} & TRKM-C$^{\dagger}$ \\ 
Dataset $\downarrow$ &    ACC (\%)  &  ACC (\%)  &  ACC (\%) &  ACC (\%) &  ACC (\%)  &   ACC (\%)    \\ 
$(\#Samples \times \#Feature)$  &   $(d_1, \sigma)$  &   $(d_1, d_2, \sigma)$  &   $(d_1, d_2, \tau, \sigma)$  &  $(\gamma, \eta, \sigma)$ &  $(d_1, d_2, \sigma)$  &   $(\gamma_1, \eta_1, \sigma)$   \\ \hline
aus & $56.25$ & $81.25$ & $79.33$ & $88.94$ & $87.98$ & $87.5$ \\
$(690 \times 15)$  & $(10^{-5}, 2^{-5})$ & $(10^{-5}, 10^{-5}, 2^{-5})$ & $(10^{2}, 10^{4}, 0.1, 2^{3})$ & $(10^{1}, 10^{3}, 2^{4})$ & $(10^{-1}, 1, 2^{5})$ & $(10^{-1}, 10^{-3}, 2^{2})$ \\
breast\_cancer & $74.42$ & $67.44$ & $63.95$ & $75.58$ & $62.79$ & $77.91$ \\
$(286 \times 10)$ & $(10^{-5}, 2^{-5})$ & $(10^{-1}, 10^{-2}, 2^{-4})$ & $(10^{2}, 10^{-2}, 1, 2^{-4})$ & $(10^{1}, 10^{-5}, 2^{4})$ & $(10^{1}, 10^{1}, 2^{5})$ & $(10^{-1}, 10^{-1}, 2^{2})$ \\
checkerboard\_Data & $56.25$ & $81.25$ & $79.33$ & $86.94$ & $87.98$ & $87.5$ \\
$(690 \times 15)$ & $(10^{-5}, 2^{-5})$ & $(10^{-2}, 10^{-2}, 2^{-5})$  &  $(10^{-1}, 10^{-5}, 0.5, 2^{-3})$  & $(10^{1}, 10^{3}, 2^{4})$ & $(10^{-1}, 1, 2^{5})$  & $(10^{-1}, 10^{-3}, 2^{2})$ \\
chess\_krvkp & $52.35$ & $90.41$ & $90.41$ & $98.27$ & $97.08$ & $98.75$ \\
$(3196 \times 37)$ & $(10^{-3}, 2^{3})$ & $(10^{-1}, 10^{-1}, 2^{-5})$  & $(10^{-5}, 10^{-5}, 0.1, 2^{4})$ & $(10^{-1}, 10^{-5}, 2^{5})$ & $(1, 1, 2^{5})$ &  $(10^{-2}, 10^{-5}, 2^{5})$ \\
crossplane130  & $51.28$ & $100$ & $97.44$ & $100$ & $100$ & $100$ \\
$(130 \times 3)$ & $(10^{-3}, 2^{4})$ & $(10^{-5}, 10^{-5}, 2^{-5})$  & $(10^{-1}, 10^{-5}, 1, 2^{-2})$  & $(10^{-5}, 10^{4}, 2^{-3})$ & $(10^{-5}, 10^{-5}, 2^{5})$ & $(10^{-5}, 10^{-5}, 2^{-5})$  \\
ecoli-0-1\_vs\_5 & $94.44$ & $97.22$ & $95.83$ & $96.81$ & $98.89$ & $97.74$ \\
$(240 \times 7)$ & $(1, 1)$ & $(10^{-5}, 10^{-4}, 2^{-3})$  &  $(10^{-3}, 10^{-5}, 0.75, 2^{4})$ & $(10^{2}, 10^{-5}, 2^{1})$  & $(10^{-4}, 10^{-5}, 2^{5})$ & $(10^{-1}, 10^{-5}, 2^{2})$ \\
ecoli-0-1\_vs\_2-3-5 & $91.89$ & $90.59$ & $94.59$ & $94.59$ & $90.81$ & $93.24$ \\
$(244 \times 8)$ & $(1, 1)$ &  $(10^{-4}, 10^{-3}, 2^{5})$ & $(10^{-5}, 10^{-5}, 0.1, 2^{3})$  &  $(10^{2}, 10^{2}, 2^{4})$ & $(10^{-2}, 1, 2^{2})$  &  $(10^{-5}, 10^{4}, 2^{-2})$ \\
ecoli-0-1-4-7\_vs\_2-3-5-6 & $87.13$ & $96.04$ & $95.05$ & $93.55$ & $88.12$ & $96.67$ \\
$(336 \times 8)$ & $(10^{-5}, 2^{-5})$ & $(10^{-1}, 1, 2^{-1})$  & $(10^{-5}, 10^{-5}, 0.1, 2^{2})$  & $(10^{-4}, 10^{2}, 2^{-1})$ & $(1, 10^{-1}, 2^{2})$ & $(10^{-5}, 10^{-5}, 2^{-5})$ \\
ecoli-0-1-4-7\_vs\_5-6 & $91$ & $96$ & $96$ & $98.36$ & $94$ & $98.46$ \\
$(332 \times 7)$ & $(10^{1}, 1)$ & $(10^{-3}, 10^{-2}, 2^{2})$  & $(10^{-1}, 10^{-5}, 0.5, 2^{-3})$  & $(10^{-5}, 10^{-5}, 1)$ & $(1, 1, 2^{5})$  & $(10^{-5}, 10^{-5}, 2^{-5})$  \\
ecoli-0-1-4-6\_vs\_5 & $98.81$ & $100$ & $100$ & $98.81$ & $97.62$ & $98.81$ \\
$(280 \times 7)$ & $(1, 1)$ &  $(10^{-5}, 10^{-5}, 2^{-5})$  & $(10^{-5}, 10^{-5}, 0.1, 2^{-3})$  & $(10^{-5}, 10^{1}, 2^{-1})$ & $(1, 1, 2^{5})$ & $(10^{-5}, 10^{-5}, 2^{-1})$ \\
haber & $82.61$ & $75.35$ & $69.57$ & $76.09$ & $77.17$ & $80.43$ \\
$(306 \times 4)$ & $(10^{-5}, 2^{-5})$ &  $(1, 10^{-1}, 2^{-3})$  & $(10^{-1}, 10^{1}, 1, 2^{-3})$ & $(10^{-1}, 10^{1}, 1)$ & $(10^{1}, 1, 2^{3})$  & $(10^{-4}, 10^{-1}, 2^{3})$ \\
haberman & $81.52$ & $75.35$ & $69.57$ & $76.09$ & $77.17$ & $80.43$ \\
$(306 \times 4)$ & $(10^{-1}, 1)$ &  $(10^{-1}, 1, 2^{-2})$ & $(10^{1}, 10^{-1}, 1, 2^{-5})$  & $(10^{-1}, 10^{1}, 1)$ & $(1, 10^{1}, 2^{2})$ & $(10^{-3}, 10^{-1}, 2^{3})$ \\
haberman\_survival & $82.61$ & $79.35$ & $69.57$ & $76.09$ & $79.35$ & $80.43$ \\
$(306 \times 4)$ & $(10^{-5}, 2^{-5})$ &  $(1, 10^{-1}, 2^{-5})$ & $(10^{-1}, 10^{1}, 1, 2^{5})$ &  $(10^{-1}, 10^{1}, 1)$ & $(10^{1}, 10^{1}, 2^{5})$ & $(10^{-4}, 10^{-1}, 2^{3})$ \\
heart-stat & $56.79$ & $70.37$ & $70.37$ & $87.65$ & $79.01$ & $85.19$ \\
$(270 \times 14)$ & $(10^{-5}, 2^{-5})$ & $(10^{-1}, 10^{-1}, 2^{-5})$  & $(10^{-5}, 1, 0.25, 2^{-5})$ & $(10^{1}, 10^{3}, 2^{4})$ & $(10^{-5}, 10^{-5}, 2^{5})$ &  $(10^{-2}, 10^{-2}, 2^{3})$ \\
led7digit-0-2-4-5-6-7-8-9\_vs\_1 & $81.95$ & $93.98$ & $93.98$ & $94.74$ & $94.74$ & $94.98$ \\
$(443 \times 8)$ & $(1, 1)$ &  $(10^{-5}, 10^{-4}, 2^{-3})$ &  $(10^{-5}, 1, 0.1, 2^{-5})$ & $(10^{1}, 10^{-5}, 2^{2})$ & $(10^{-5}, 10^{-5}, 2^{5})$  & $(10^{-2}, 10^{-4}, 2^{-5})$ \\
mammographic & $52.94$ & $79.93$ & $77.82$ & $78.74$ & $81.31$ & $79.24$ \\
$(961 \times 6)$ & $(10^{-5}, 2^{-5})$ & $(10^{-1}, 10^{-1}, 2^{-5})$  & $(10^{3}, 10^{-5}, 1, 2^{-5})$ & $(10^{-1}, 10^{-5}, 2^{4})$ & $(10^{-1}, 10^{-1}, 2^{5})$ & $(10^{-5}, 10^{-2}, 2^{-2})$ \\
monks\_3 & $46.11$ & $75.21$ & $95.81$ & $49.1$ & $80.24$ & $80.84$ \\
$(554 \times 7)$ & $(10^{-5}, 2^{-5})$ &  $(1, 10^{-1}, 2^{-2})$ & $(10^{1}, 1, 0.25, 2^{-3})$ & $(10^{5}, 10^{2}, 2^{-5})$ & $(1, 10^{1}, 2^{5})$ & $(10^{2}, 10^{-4}, 2^{-2})$ \\
mushroom & $63.41$ & $70.65$ & $80.51$ & $81.77$ & $84.91$ & $96.41$ \\
$(8124 \times 22)$ & $(10^{-3}, 2^{1})$  & $(10^{2}, 10^{3}, 2^{5})$  & $(10^{1}, 10^{2}, 0.25, 2^{-3})$ & $(10^{-1}, 10^{-5}, 2^{1})$ & $(1, 10^{-1}, 2^{3})$ & $(10^{-4}, 10^{-2}, 2^{5})$ \\
musk\_1 & $53.15$ & $83.15$ & $53.15$ & $92.31$ & $91.61$ & $92.31$ \\
$(476 \times 167)$ & $(10^{-5}, 2^{-5})$ &  $(10^{-5}, 10^{-5}, 2^{-5})$ & $(10^{-5}, 10^{-5}, 0.1, 2^{-1})$ & $(10^{1}, 10^{-5}, 2^{5})$ & $(10^{-1}, 1, 2^{5})$ & $(10^{-1}, 10^{-5}, 2^{5})$ \\
new-thyroid1 & $87.69$ & $98.46$ & $96.92$ & $98.46$ & $95.38$ & $96.92$ \\
$(215 \times 16)$ & $(10^{-5}, 2^{-5})$ & $(10^{-2}, 10^{-2}, 2^{-3})$ & $(1, 10^{1}, 0.75, 2^{-5})$ & $(1, 10^{3}, 2^{1})$ & $(10^{-5}, 10^{-5}, 2^{4})$ & $(10^{-2}, 10^{-5}, 2^{3})$ \\
oocytes\_merluccius\_nucleus\_4d & $64.82$ & $76.22$ & $79.48$ & $75.06$ & $77.2$ & $79.15$ \\
$(1022 \times 42)$ & $(10^{-5}, 2^{-5})$ & $(10^{-2}, 10^{-1}, 2^{-5})$  & $(1, 1, 0.25, 2^{-4})$ & $(10^{-1}, 10^{-1}, 2^{5})$ & $(1, 1, 2^{4})$  & $(10^{-1}, 10^{-2}, 2^{2})$ \\
ozone & $96.58$ & $96.58$ & $96.58$ & $87.96$ & $94.09$ & $97.58$ \\
$(2536 \times 6)$ & $(10^{-5}, 2^{-5})$ & $(10^{-5}, 10^{-5}, 2^{-5})$   & $(10^{-5}, 10^{-5}, 0.1, 2^{-3})$ & $(10^{-1}, 1, 2^{5})$ & $(10^{1}, 1, 2^{5})$ & $(10^{-2}, 10^{-2}, 2^{3})$  \\
ringnorm & $90.42$ & $92.65$ & $93.06$ & $89.07$ & $96.94$ & $96.44$ \\
 $(7400 \times 21)$ & $(1, 2^{1})$ &  $(10^{-5}, 10^{-5}, 2^{-5})$ & $(10^{-5}, 10^{-5}, 0.1, 2^{-3})$ & $(10^{-2}, 10^{-5}, 2^{1})$ & $(10^{1}, 1, 2^{5})$ & $(10^{-5}, 10^{-5}, 1)$  \\
shuttle-6\_vs\_2-3 & $95.65$ & $97.1$ & $97.1$ & $100$ & $98.55$ & $98.55$ \\
$(230 \times 10)$ & $(10^{-5}, 2^{-5})$ & $(10^{-5}, 10^{-4}, 2^{-4})$   & $(10^{-5}, 10^{-5}, 0.1, 2^{-4})$ & $(10^{-5}, 10^{5}, 2^{3})$ & $(10^{-2}, 10^{-1}, 2^{4})$ & $(10^{-5}, 10^{4}, 1)$ \\
spambase & $62.2$ & $84.79$ & $76.76$ & $79.37$ & $90.88$ & $98$ \\
$(4601 \times 58)$ & $(10^{-1}, 1)$ & $(10^{-2}, 1, 2^{-3})$  &  $(10^{3}, 10^{4}, 0.5, 2^{3})$ & $(10^{2}, 10^{2}, 2^{4})$ & $(10^{1}, 10^{1}, 2^{5})$ & $(10^{-5}, 10^{3}, 2^{2})$ \\
spectf & $80.25$ & $79.42$ & $83.95$ & $83.95$ & $85.19$ & $85.19$ \\
$(267 \times 45)$ & $(10^{-5}, 2^{-5})$ & $(10^{-1}, 10^{-1}, 2^{-5})$  & $(10^{-5}, 10^{-4}, 0.75, 2^{-3})$ & $(10^{1}, 1, 2^{3})$ & $(10^{-1}, 10^{-3}, 2^{4})$ & $(10^{-4}, 10^{-5}, 2^{1})$ \\
tic\_tac\_toe & $66.32$ & $95$ & $100$ & $100$ & $98.96$ & $99.65$ \\
$(958 \times 10)$ & $(10^{-5}, 2^{-5})$ & $(10^{-1}, 10^{-2}, 2^{-2})$   & $(10^{-5}, 10^{-5}, 0.1, 2^{2})$   & $(10^{-5}, 10^{3}, 2^{-2})$ & $(10^{1}, 10^{1}, 2^{4})$ & $(10^{-1}, 10^{-5}, 2^{5})$  \\
vehicle1 & $75.98$ & $80.31$ & $77.95$ & $88.58$ & $81.5$ & $81.89$ \\
$(846 \times 19)$ & $(10^{-5}, 2^{-5})$ & $(10^{-2}, 10^{-2}, 2^{-5})$ & $(10^{-2}, 1, 0.5, 2^{3})$ & $(10^{-1}, 1, 2^{5})$ & $(1, 1, 2^{5})$ & $(10^{-3}, 1, 2^{2})$ \\
vehicle2 & $77.95$ & $76.46$ & $96.06$ & $98.82$ & $94.49$ & $95.28$ \\
$(846 \times 19)$ & $(1, 1)$ & $(10^{-2}, 10^{-1}, 2^{-2})$ & $(10^{-2}, 1, 0.5, 2^{5})$ & $(10^{-1}, 10^{-1}, 2^{5})$ & $(10^{-1}, 10^{-1}, 2^{5})$ & $(10^{-5}, 10^{-5}, 1)$ \\
vertebral\_column\_2clases & $69.89$ & $89.25$ & $81.72$ & $89.25$ & $88.17$ & $87.1$ \\
$(310 \times 7)$ & $(10^{-5}, 2^{-5})$ & $(10^{-2}, 10^{-2}, 2^{-5})$  & $(1, 1, 0.25, 2^{-3})$ & $(1, 10^{5}, 2^{5})$ & $(10^{-1}, 10^{-1}, 2^{4})$ & $(10^{-3}, 1, 2^{1})$ \\
wpbc & $77.97$ & $77.97$ & $77.97$ & $77.97$ & $76.27$ & $76.27$ \\
$(194 \times 34)$ & $(10^{-5}, 2^{-5})$ & $(10^{-5}, 10^{-5}, 2^{-5})$ & $(10^{-5}, 10^{-5}, 0.1, 2^{5})$ & $(10^{1}, 10^{-1}, 2^{4})$ & $(10^{-5}, 10^{-5}, 2^{5})$ & $(10^{-3}, 1, 2^{4})$ \\
yeast-0-2-5-6\_vs\_3-7-8-9 & $91.39$ & $84.04$ & $94.37$ & $94.04$ & $94.04$ & $92.72$ \\
$(1004 \times 9)$ & $(10^{-5}, 2^{-5})$ & $(10^{-1}, 10^{-1}, 2^{1})$ & $(10^{-1}, 10^{1}, 0.75, 2^{-2})$  & $(10^{2}, 10^{3}, 2^{4})$ & $(1, 1, 2^{5})$  & $(10^{-5}, 10^{3}, 2^{-1})$ \\
yeast-0-3-5-9\_vs\_7-8 & $87.5$ & $53.15$ & $95.7$ & $97.35$ & $90.79$ & $95.7$ \\
$(1004 \times 9)$ & $(10^{-5}, 2^{-5})$ & $(10^{-1}, 10^{-5}, 2^{-5})$ & $(10^{2}, 10^{3}, 0.1, 2^{2})$ & $(10^{3}, 10^{4}, 2^{5})$ & $(10^{-5}, 10^{-5}, 2^{5})$  &  $(1, 10^{-2}, 2^{1})$\\
yeast-0-5-6-7-9\_vs\_4 & $91.19$ & $82.45$ & $88.68$ & $91.82$ & $91.19$ & $92.45$ \\
$(528 \times 9)$ & $(10^{-5}, 2^{-5})$ &  $(10^{-3}, 10^{-2}, 2^{-4})$  & $(10^{5}, 10^{5}, 0.1, 2^{5})$ & $(10^{1}, 1, 2^{2})$ & $(10^{-2}, 10^{-2}, 2^{5})$ & $(10^{-5}, 10^{2}, 2^{-1})$ \\
yeast-2\_vs\_4 & $85.81$ & $94.19$ & $94.19$ & $96.77$ & $97.42$ & $97.32$ \\
$(514 \times 9)$ & $(10^{-5}, 2^{-5})$ & $(10^{-1}, 10^{-1}, 2^{-5})$ & $(10^{-2}, 1, 0.5, 2^{2})$ & $(10^{1}, 10^{-1}, 2^{3})$ & $(10^{-1}, 10^{-1}, 2^{5})$ & $(10^{-4}, 10^{2}, 1)$ \\
yeast3 & $89.24$ & $92.38$ & $91.26$ & $93.95$ & $92.83$ & $93.5$ \\ 
$(1484 \times 9)$ & $(10^{-5}, 2^{-5})$  &  $(10^{-2}, 10^{-1}, 2^{-3})$ & $(10^{3}, 1, 0.5, 2^{4})$  & $(10^{1}, 10^{-1}, 2^{2})$ & $(1, 10^{2}, 2^{5})$ & $(10^{-5}, 10^{-5}, 2^{-1})$ \\ \hline
Avergae ACC & $76.27$ & $84.83$ & $85.95$ & $88.52$ & $\underline{88.74}$ & $\textbf{90.85}$ \\ \hline
Avergae Rank & $4.98$ & $4$ & $3.86$ & $2.77$ & $3.16$ & $2.20$ \\  \hline
\multicolumn{7}{l}{$^{\dagger}$ represents the proposed model.} \\
\multicolumn{7}{l}{The top and second-best models in terms of ACC are denoted by boldface and underline, respectively.}
\end{tabular}}
\end{table*}

In this subsection, we offer a thorough comparison of the proposed TRKM-C model against the SVM \cite{cortes1995support}, TSVM \cite{khemchandani2007twin}, Pin-GTSVM \cite{tanveer2019general}, RKM \cite{suykens2017deep}, and GBTSVM \cite{GBquadir2024granular} models. $36$ benchmark datasets from the KEEL and UCI repositories are used in this comparison. The best hypermeter and the average classification accuracy (ACC) of the proposed TRKM-C, along with the baseline SVM, TSVM, Pin-GTSVM, RKM, and GBTSVM models, are presented in Table \ref{Average ACC and average rank for UCI and KEEL datasets}. The ACC comparison demonstrates that our proposed TRKM-C model outperforms the baseline SVM, TSVM, Pin-GTSVM, RKM, and GBTSVM models on most of the datasets. On several datasets, the proposed TRKM-C demonstrates clear improvements over baseline models. For instance, on the chess\_krvkp dataset, TRKM-C achieves an ACC of $98.75\%$, surpassing SVM ($52.35\%$), TSVM ($90.4\%$), and RKM ($98.27\%$). Similarly, for the mushroom dataset, TRKM-C attains $96.41\%$ ACC, outperforming TSVM ($70.65\%$), Pin-GTSVM ($80.51\%$), and GBTSVM ($84.91\%$). On smaller datasets like breast\_cancer, TRKM-C also shows superior performance with $77.91\%$ ACC, compared to SVM ($74.42\%$) and RKM ($75.58\%$). Furthermore, for medium-sized datasets such as ecoli-0-1-4-7\_vs\_2-3-5-6 and  ecoli-0-1-4-7\_vs\_5-6, TRKM-C maintains consistent superiority, achieving $96.67\%$ and $98.46\%$ ACC, respectively, while SVM and TSVM record lower ACCs. The proposed TRKM-C model provides an average ACC of $90.85\%$,  whereas the average ACCs of SVM, TSVM, Pin-GTSVM, RKM, and GBTSVM models are $76.27\%$, $84.83\%$, $85.95\%$, $88.52\%$, and $88.74$, respectively. The proposed TRKM-C model achieves a higher average ACC compared to the existing models. Average accuracy can sometimes be misleading if a model performs exceptionally well on one dataset but poorly across others. This can skew the results and may not provide a complete picture of the model's performance. To gauge the effectiveness and performance of the models, we employ the ranking method outlined by \citet{demvsar2006statistical}. In this method, classifiers are ranked such that models with better performance receive lower ranks, while those with poorer performance are assigned higher ranks. The rank of the \( j^{th} \) model on the \( i^{th} \) dataset is expressed as \( \mathfrak{R}_{j}^{i} \) in order to evaluate \( p \) models over \( N \) datasets. The average rank of the model is determined as \( \mathfrak{R}_j = \frac{1}{N} \sum_{i=1}^{N} \mathfrak{R}_{j}^i \). The average rank of the proposed TRKM-C, along with the baselines SVM, TSVM, Pin-GTSVM, RKM, and GBTSVM models, are $2.20$, $4.98$, $4.00$, $3.86$, $2.77$, and $3.16$, respectively. The lowest average rank is attained by the proposed TRKM-C model, indicating the most favorable position compared to the existing models. The Friedman test \cite{demvsar2006statistical} is then used to determine whether the models differ significantly from one another. Under the null hypothesis of the Friedman test, it is assumed that all models have the same average rank, indicating equivalent performance. The Friedman test follows a chi-squared distribution (\( \chi^2_F \)) with \( (p - 1) \) degrees of freedom. Its calculation is given by:  $\chi^2_F = \frac{12N}{p(p+1)}\left[\sum_{j}\mathfrak{R}_j^2 - \frac{p(p+1)^2}{4}\right].$ The $F_F$ statistic is calculated using the formula: $F_F = \frac{(N - 1) \chi^2_F}{N(p - 1) - \chi^2_F}$, where the $F$-distribution has \( (p - 1) \) and \( (N - 1) \times (p - 1) \) degrees of freedom. For \( p = 6 \) and \( N = 36 \), we obtain \( \chi^2_F = 48.95 \) and \( F_F = 13.0732 \) at the $5\%$ significance level. According to the statistical $F$-distribution table, \( F_F(5, 175) = 2.2657 \). Since $F_F > 2.2657$, the null hypothesis is rejected. As a result, there is a statistically significant disparity among the models under comparison. The Nemenyi post hoc test is then used to assess the models' pairwise differences. The critical difference (C.D.) is calculated using the formula \(\text{C.D.} = q_\alpha \sqrt{\frac{p(p + 1)}{6N}}\), where \( q_\alpha \) is the two-tailed Nemenyi test critical value obtained from the distribution table. With reference to the $F$-distribution table, the computed C.D. is $1.2567$ at the $5\%$ significance level, with \(q_\alpha = 2.850 \). The proposed TRKM-C model's average rank differences with the SVM, TSVM, Pin-GTSVM, RKM, and GBTSVM models are as follows: $2.78$, $1.80$, $1.66$, $0.57$, and $0.96$, respectively. The Nemenyi post hoc test confirms that the proposed TRKM-C model is significantly superior to the baseline SVM, TSVM, and Pin-GTSVM models. The proposed TRKM-C model shows a statistically significant improvement over the RKM and GBTSVM models. Based on the lowest average rank attained by the TRKM-C model, we conclude that the proposed TRKM-C model excels in overall performance and ranking compared to existing models.
\begin{table}[htp]
\centering
    \caption{Comparison of Win-Tie-Loss Results on UCI and KEEL classification datasets.}
    \label{win tie loss sign test}
    \resizebox{0.9\textwidth}{!}{
    % \resizebox{\columnwidth}{!}{%
\begin{tabular}{lccccc}
\hline
\textbf{}   & SVM \cite{cortes1995support}         &  TSVM \cite{khemchandani2007twin}    & Pin-GTSVM \cite{tanveer2019general}  & RKM \cite{suykens2017deep} & GBTSVM \cite{GBquadir2024granular} \\ \hline
TSVM \cite{khemchandani2007twin} & {[}$24, 2, 10${]} &  &  &  &  \\
Pin-GTSVM \cite{tanveer2019general} & {[}$28, 3, 5${]} & {[}$11, 9, 16${]} &  &  &  \\
RKM \cite{suykens2017deep} & {[}$29, 2, 5${]} & {[}$22, 4, 10${]} & {[}$25, 4, 7${]} &  &  \\
GBTSVM \cite{GBquadir2024granular} & {[}$27, 1, 8${]} & {[}$26, 2, 8${]} & {[}$22, 0, 14${]} & {[}$14, 3, 19${]} &  \\
TRKM-C$^{\dagger}$ & {[}$31, 1, 4${]} & {[}$30, 1, 5${]} & {[}$26, 2, 8${]} & {[}$20, 3, 13${]} & {[}$24, 4, 8${]} \\ \hline
\multicolumn{6}{l}{$^{\dagger}$ represents the proposed model.}
\end{tabular}}
\end{table}
Furthermore, to evaluate the model's performance, we use the pairwise win-tie-loss sign test. Under the null hypothesis, the test assumes that two models perform equally well, each expected to be the best in half of the $N$ datasets. A model is considered significantly superior if it surpasses the competition on approximately \( N/2 + 1.96 \sqrt{N/2} \) datasets. The two models are distributed equally if the number of ties between them is even. One tie is eliminated, and the remaining ties are divided equally among the classifiers if the number of ties is odd. For \( N = 36 \), a model needs to secure at least $24$ wins to demonstrate a significant difference from the other models. The proposed TRKM-C model is evaluated in comparison to the baseline models in Table \ref{win tie loss sign test}. The entry $[x, y, z]$ in Table \ref{win tie loss sign test} shows that, when compared to the model listed in the column, the model listed in the row wins $x$ times, ties $y$ times, and loses $z$ times. The proposed TRKM-C model demonstrates a statistically significant difference compared to the baseline models, with the exception of RKM. The TRKM-C model consistently demonstrates its superior performance over the RKM model, as evidenced by its higher winning percentage. The results indicate that the proposed TRKM-C model considerably surpasses the performance of the baseline models.

\begin{table*}[htp]
\centering
    \caption{Testing ACC and training time of classifiers on real-world datasets.}
    \label{Testing ACC and training time of classifiers on real-world datasets.}
    \resizebox{0.9\linewidth}{!}{
\begin{tabular}{lcccccc}
\hline
Model $\rightarrow$ & SVM \cite{cortes1995support} & TSVM \cite{khemchandani2007twin} & Pin-GTSVM \cite{tanveer2019general} & RKM \cite{suykens2017deep} & GBTSVM \cite{GBquadir2024granular} & TRKM-C$^{\dagger}$ \\ 
Dataset $\downarrow$ &    ACC (\%)  &  ACC (\%)  &  ACC (\%) &  ACC (\%) &  ACC (\%)  &   ACC (\%)    \\ 
 & Time(s) & Time(s) & Time(s) & Time(s) & Time(s) & Time(s) \\ \hline
adult & 80.52 & 81.23 & 81.89 & 80.36 & 82.48 & 83.27 \\
$(48842 \times15)$ & 600.74 & 1689.28 & 649.77 & 482.71 & 209.35 & 243.69 \\
connect\_4 & b & b & b & 77.25 & 77.49 & 78.46 \\
$(67557 \times43)$ &  &  &  & 587.42 & 256.32 & 285.47 \\
creditcard & b & b & b & 74.78 & 75.42 & 75.36 \\
$(284807 \times30)$ &  &  &  & 709.82 & 298.63 & 343.28 \\
miniboone & b & b & b & 85.89 & 85.42 & 86.43 \\
$(130064 \times51)$ &  &  &  & 668.29 & 298.36 & 323.65 \\
\hline
\multicolumn{7}{l}{\begin{tabular}[c]{@{}l@{}} $^b$ Terminated because of out of memory.\end{tabular}} \\
\multicolumn{7}{l}{$^{\dagger}$ represents the proposed model.}
\end{tabular}}
\end{table*}

\subsection{Experimental Validation on Large-Scale Datasets}
In this subsection, we have conducted the experiments on multiple large-scale real-world datasets, including adult, Connect\_4, creditcard, and miniboone, to demonstrate TRKM-C’s computational robustness and efficiency in handling high-dimensional and large-sample problems. These datasets contain up to $\approx 284807$ samples and over $50$ features, effectively validating the model’s applicability beyond small or medium-scale settings.
Table \ref{Testing ACC and training time of classifiers on real-world datasets.} reports the ACC and training time (in seconds) of TRKM-C compared with baseline models. 
\\
First, in terms of classification ACC, TRKM-C consistently achieves the highest or near-highest performance across all datasets. For instance, on the adult and miniboone datasets, TRKM-C attains $83.27\%$ and $86.43\%$, respectively, surpassing RKM and GBTSVM by notable margins. These results confirm that the conjugate feature representation in TRKM-C effectively captures discriminative structures in large-scale data without overfitting, maintaining robust generalization performance.
\\
Second, regarding computational efficiency, TRKM-C exhibits remarkable scalability. While traditional models such as SVM, TSVM, and Pin-GTSVM terminate due to memory exhaustion on large datasets (as indicated by the superscript $b$), TRKM-C completes all runs successfully with training times significantly lower than RKM and comparable to or better than GBTSVM. This efficiency stems from the dual formulation derived via Fenchel-Young conjugacy, which transforms the optimization into two smaller and independent subproblems, thus reducing computational and memory overhead.
\\
These results provide strong empirical evidence that TRKM-C not only maintains superior classification accuracy but also scales efficiently to real-world, high-volume datasets where traditional kernel-based approaches fail. The combination of its energy-based conjugate formulation and twin-structured decomposition enables it to achieve a favorable trade-off between accuracy and computational cost, thus fulfilling the original motivation of addressing the computational bottlenecks inherent in large-scale kernel learning.

\subsection{Experiment on Artificial NDC Datasets}
To evaluate the performance and scalability of the proposed TRKM model, we conducted experiments on the NDC datasets, which are designed to challenge machine learning models with large-scale data ranging from 10k (10,000 samples) to 5l (500,000 samples), each with a fixed feature dimension of 32. The results, presented in Table \ref{Testing ACC and training time of classifiers on NDC datasets}, compare TRKM’s testing accuracy (ACC \%) and training time (in seconds) against baseline models, including SVM, TSVM, Pin-GTSVM, RKM, and GBTSVM. TRKM consistently achieves the highest accuracies across all datasets, ranging from $83.92\%$ on NDC-10k to $86.36\%$ on NDC-5l, surpassing GBTSVM ($83.89\%$ to $85.89\%$) and RKM ($79.49\%$ to $84.67\%$), particularly on larger datasets where its robustness in handling complex data is evident. In terms of training time, TRKM demonstrates competitive efficiency, with times ranging from $96.76$ seconds (NDC-10k) to $518.67$ seconds (NDC-5l), slightly higher than GBTSVM in some cases (e.g., $475.69$ seconds on NDC-5l) but significantly faster than SVM (up to $815.36$ seconds on NDC-50k) and TSVM (up to $2569.32$ seconds on NDC-50k). Compared to RKM, TRKM reduces computational overhead, with training times such as $301.58$ seconds versus $659.82$ seconds on NDC-1l. Notably, baseline models like SVM, TSVM, and Pin-GTSVM encountered ``out-of-memory'' issues on datasets of 1l and larger, while TRKM and GBTSVM scaled effectively, with TRKM maintaining superior accuracy. The efficiency of TRKM stems from its integration of the kernel trick, conjugate feature duality based on the Fenchel-Young inequality, and a regularized least squares approach, which collectively mitigate the computational bottlenecks of RKM while leveraging the strengths of twin methods. These results underscore TRKM’s ability to deliver high accuracy and scalability, making it a robust solution for large-scale classification tasks and a significant advancement over existing kernel-based and twin support vector machine models.

\begin{table*}[htp]
\centering
    \caption{Testing ACC and training time of classifiers on NDC datasets.}
    \label{Testing ACC and training time of classifiers on NDC datasets}
    \resizebox{0.9\linewidth}{!}{
\begin{tabular}{lcccccc}
\hline
Model $\rightarrow$ & SVM \cite{cortes1995support} & TSVM \cite{khemchandani2007twin} & Pin-GTSVM \cite{tanveer2019general} & RKM \cite{suykens2017deep} & GBTSVM \cite{GBquadir2024granular} & TRKM-C$^{\dagger}$ \\ 
Dataset $\downarrow$ &    ACC (\%)  &  ACC (\%)  &  ACC (\%) &  ACC (\%) &  ACC (\%)  &   ACC (\%)    \\ 
 & Time(s) & Time(s) & Time(s) & Time(s) & Time(s) & Time(s) \\ \hline
NDC-10k & \begin{tabular}[c]{@{}c@{}}81.64\\ 312.65\end{tabular} & \begin{tabular}[c]{@{}c@{}}80.43\\ 1256.85\end{tabular} & \begin{tabular}[c]{@{}c@{}}81.28\\ 216.72\end{tabular} & \begin{tabular}[c]{@{}c@{}}82.84\\ 114.72\end{tabular} & \begin{tabular}[c]{@{}c@{}}83.92\\ 94.85\end{tabular} & \begin{tabular}[c]{@{}c@{}}83.89\\ 96.76\end{tabular} \\
NDC-50k & \begin{tabular}[c]{@{}c@{}}75.82\\ 815.36\end{tabular} & \begin{tabular}[c]{@{}c@{}}76.82\\ 2569.32\end{tabular} & \begin{tabular}[c]{@{}c@{}}77.22\\ 736.89\end{tabular} & \begin{tabular}[c]{@{}c@{}}79.49\\ 529.28\end{tabular} & \begin{tabular}[c]{@{}c@{}}81.42\\ 228.46\end{tabular} & \begin{tabular}[c]{@{}c@{}}82.69\\ 259.45\end{tabular} \\
NDC-1l & $^b$ & $^b$ & $^b$ & \begin{tabular}[c]{@{}c@{}}84.67\\ 659.82\end{tabular} & \begin{tabular}[c]{@{}c@{}}84.89\\ 287.36\end{tabular} & \begin{tabular}[c]{@{}c@{}}85.26\\ 301.58\end{tabular} \\
NDC-3l & $^b$ & $^b$ & $^b$ & \begin{tabular}[c]{@{}c@{}}80.24\\ 723.58\end{tabular} & \begin{tabular}[c]{@{}c@{}}81.23\\ 310.59\end{tabular} & \begin{tabular}[c]{@{}c@{}}81.56\\ 355.48\end{tabular} \\
NDC-5l & $^b$ & $^b$ & $^b$ & $^b$ & \begin{tabular}[c]{@{}c@{}}85.89\\ 475.69\end{tabular} & \begin{tabular}[c]{@{}c@{}}86.36\\ 518.67\end{tabular} \\ \hline
\multicolumn{7}{l}{\begin{tabular}[c]{@{}l@{}} $^b$ Terminated because of out of memory.\end{tabular}} \\
\multicolumn{7}{l}{$^{\dagger}$ represents the proposed model.}
\end{tabular}}
\end{table*}

\subsection{Sensitivity Analysis on Real World UCI and KEEL Datasets for Classification}
\begin{figure*}[ht!]
\begin{minipage}{.246\linewidth}
\centering
\subfloat[haber]{\label{1a}\includegraphics[scale=0.20]{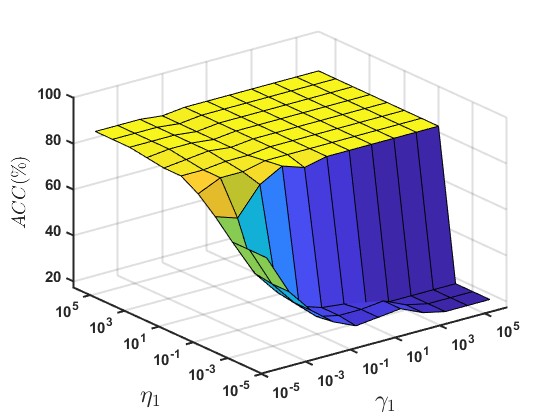}}
\end{minipage}
% \par\medskip
% \par\medskip
\begin{minipage}{.246\linewidth}
\centering
\subfloat[monks\_3]{\label{1b}\includegraphics[scale=0.20]{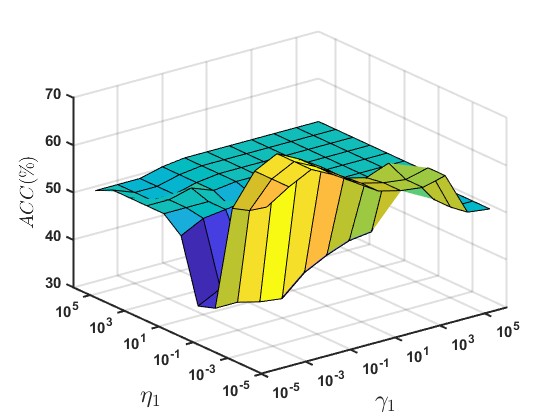}}
\end{minipage}
% \par\medskip
\begin{minipage}{.246\linewidth}
\centering
\subfloat[new-thyroid1]{\label{1c}\includegraphics[scale=0.20]{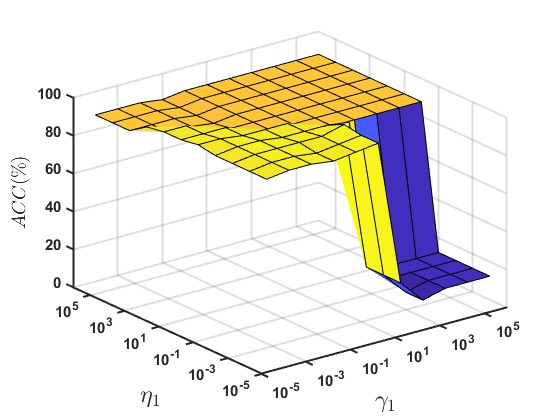}}
\end{minipage}
% \par\medskip
% \par\medskip
\begin{minipage}{.246\linewidth}
\centering
\subfloat[breast\_cancer]{\label{1d}\includegraphics[scale=0.20]{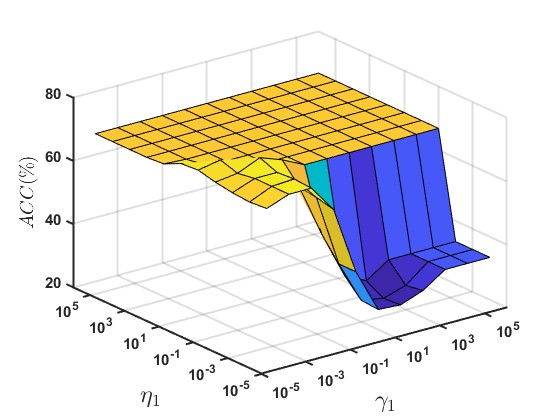}}
\end{minipage}
\caption{The impact of changing the parameters \(\eta\) and \(\gamma\) on the ACC values of the proposed TRKM-C model.}
\label{Effect of parameters eta and gamma}
\end{figure*}

\begin{figure*}[ht!]
\begin{minipage}{.246\linewidth}
\centering
\subfloat[haber]{\label{2a}\includegraphics[scale=0.20]{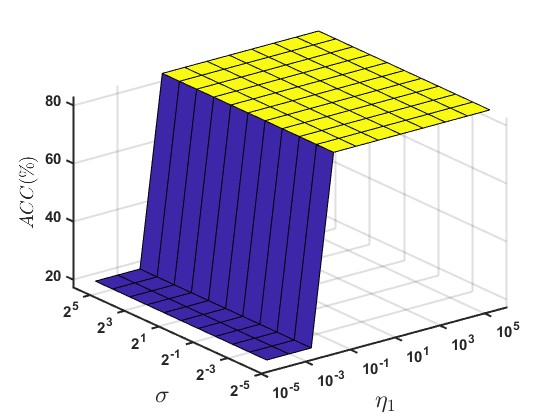}}
\end{minipage}
% \par\medskip
% \par\medskip
\begin{minipage}{.246\linewidth}
\centering
\subfloat[monks\_3]{\label{2b}\includegraphics[scale=0.20]{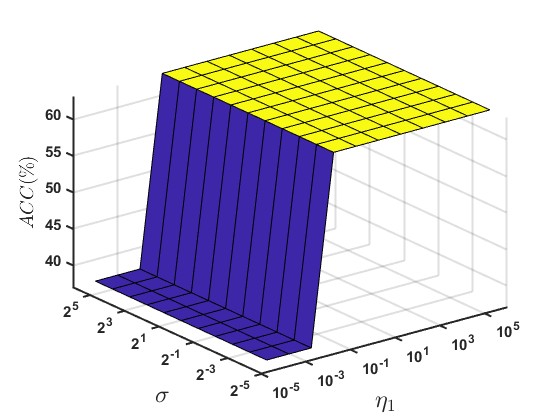}}
\end{minipage}
% \par\medskip
\begin{minipage}{.246\linewidth}
\centering
\subfloat[ringnorm]{\label{2c}\includegraphics[scale=0.20]{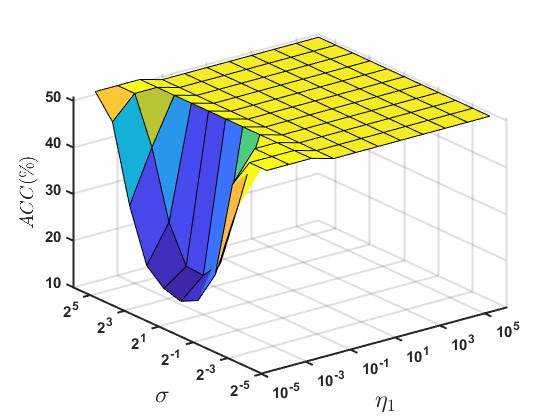}}
\end{minipage}
% \par\medskip
% \par\medskip
\begin{minipage}{.246\linewidth}
\centering
\subfloat[breast\_cancer]{\label{2d}\includegraphics[scale=0.20]{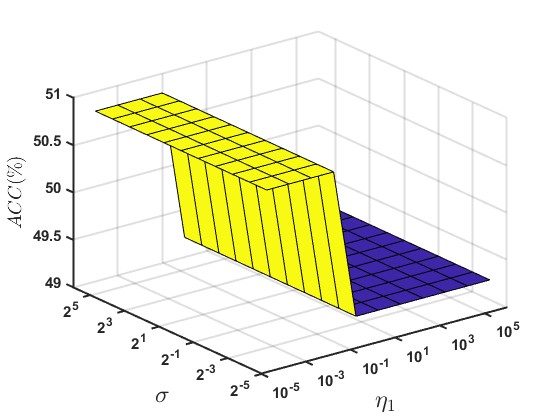}}
\end{minipage}
\caption{The impact of changing the parameters $\eta$ and $\sigma$ on the ACC values of the proposed TRKM-C model.}
\label{Effect of parameters eta and sigma}
\end{figure*}

A comprehensive evaluation of the proposed TRKM-C model's robustness requires analyzing its sensitivity to different hyperparameters. We conduct sensitivity analyses focusing on the following aspects: (1) $\eta$ versus $\gamma$, and (2) \(\eta\) versus $\sigma$. We experiment with different ranges for each hyperparameter and assess their impact on the model's performance.
\begin{enumerate}
    \item A complete assessment of the robustness of the proposed TRKM-C model requires examining its sensitivity to the hyperparameters \(\eta\) and \(\gamma\). This thorough investigation aids in determining the setup that optimizes predictive ACC and improves the model's performance on unknown inputs. Fig. \ref{Effect of parameters eta and gamma} highlights significant fluctuations in model ACC across various \(\eta\) and \(\gamma\) values, demonstrating the sensitivity of the model to these hyperparameters. From Figs. \ref{1a} and \ref{1b}, the TRKM-C model performs best inside the ranges of \(\eta\) from \(10^{-1}\) to \(10^{5}\) and \(\gamma\) from \(10^{-5}\) to \(10^{-1}\). Also, Figs. \ref{1c} and \ref{1d} show that the model achieves maximum ACC when \(\eta\) and \(\gamma\) are within \(10^{-5}\) to \(10^{3}\). Therefore, we recommend using \(\eta\) and \(\gamma\) within the range of \(10^{-5}\) to \(10^{-1}\) for optimal results. However, fine-tuning may be required depending on the specific characteristics of the dataset to achieve the best generalization performance with the TRKM-C model.
    \item We assess the performance of the proposed TRKM-C model by altering the values of \(\eta\) and \(\sigma\). Fig. \ref{Effect of parameters eta and sigma} shows significant variations in model ACC across different combinations of \(\eta\) and \(\sigma\), underscoring the sensitivity of the proposed TRKM-C model to these hyperparameters. We can see from the results shown in Figs. \ref{2a} and \ref{2b} that the proposed TRKM-C model performs well within \(\gamma\) ranges of \(10^{-3}\) to \(10^5\).  Fig \ref{2c} shows an increase in testing ACC within the \(\eta\) range of \(10^{-1}\) to \(10^{5}\) and the \(\sigma\) range of \(2^{-3}\) to \(2^{3}\). Similarly, Fig. \ref{2d} shows that testing ACC increases as \(\eta\) ranges from \(10^{-5}\) to \(10^{-1}\) across all ranges of \(\sigma\). Therefore, it is essential to carefully select the hyperparameters for the TRKM-C model to attain optimal generalization performance.
\end{enumerate}

\begin{table*}[ht!]
\centering
    \caption{RMSE, MAE, Pos Error, and Neg Error values of the proposed TRKM-R and the baseline models across regression datasets.}
    \label{Average RMSE and average rank for UCI datasets}
    \resizebox{0.9\linewidth}{!}{
\begin{tabular}{llccccc}
\hline
Dataset $\downarrow$ Model $\rightarrow$ &  & SVR \cite{basak2007support} & TSVR \cite{peng2010tsvr} & TSVQR \cite{ye2024twin} & RKM \cite{suykens2017deep} & TRKM-R$^{\dagger}$ \\ 
$(\#Samples \times \#Feature)$ &   &   &   &  &  &   \\ \hline
Abalone & RMSE & $0.00619696$ & $0.38767592$ & $0.07666791$ & $0.00044776$ & $0.00003633$ \\
$(4117 \times 7)$ & MAE & $0.00448455$ & $0.25410483$ & $6.07666478$ & $0.00000301$ & $0.00002449$ \\
 & Pos Error & $0.00327$ & $0.26615116$ & $1.20577705$ & $0.00000309$ & $0.00001841$ \\
 & Neg Error & $0.0063161$ & $0.00007567$ & $6.07666478$ & $0.00000293$ & $0.00002719$ \\ \hline
Airfoil\_Self\_Noise & RMSE & $0.28273723$ & $0.00088726$ & $0.25334222$ & $0.00033556$ & $0.00054951$ \\
$(1503 \times 5)$ & MAE & $0.20323196$ & $0.28901812$ & $0.20892536$ & $0.00000518$ & $0.00053706$ \\
 & Pos Error & $0.30709839$ & $0.22327934$ & $0.32777245$ & $0.00000503$ & $0.00053843$ \\
 & Neg Error & $0.13613576$ & $0.76480796$ & $0.17763229$ & $0.00000535$ & $0.00033233$ \\  \hline
auto-original & RMSE & $0.02208954$ & $0.33117149$ & $0.66636073$ & $0.00039196$ & $0.00026814$ \\
$(392 \times 7)$ & MAE & $0.01923475$ & $0.0577438$ & $0.52216993$ & $0.00022567$ & $0.00018836$ \\
 & Pos Error & $0.02076571$ & $0.03669568$ & $0.58268115$ & $0.00032394$ & $0.00022018$ \\
 & Neg Error & $0.00786152$ & $0.24246004$ & $0.25812095$ & $0.00082211$ & $0.00010973$ \\   \hline
Auto-price & RMSE & $0.24931086$ & $0.38767592$ & $0.80669028$ & $0.00047829$ & $0.00043308$ \\
$(159 \times 15)$ & MAE & $0.19605144$ & $0.07397557$ & $0.50818611$ & $0.00062508$ & $0.00031579$ \\
 & Pos Error & $0.2362126$ & $0.12724069$ & $0.57794927$ & $0.00018158$ & $0.00016123$ \\
 & Neg Error & $0.07556795$ & $0.32188751$ & $0.32036222$ & $0.00013476$ & $0.00038605$ \\   \hline
bodyfat & RMSE & $0.04248758$ & $0.1320877$ & $0.7909357$ & $0.0003355$ & $0.00074434$ \\
$(252 \times 14)$ & MAE & $0.03476774$ & $0.06042572$ & $0.62823812$ & $0.00084039$ & $0.00060215$ \\
 & Pos Error & $0.03072008$ & $0.08347705$ & $0.59351638$ & $0.00079833$ & $0.00064703$ \\
 & Neg Error & $0.03712887$ & $0.06934175$ & $0.65488504$ & $0.00057252$ & $0.00036282$ \\  \hline
cpu\_pref & RMSE & $0.04585881$ & $0.27085771$ & $1.40835874$ & $0.00476195$ & $0.0013733$ \\   
$(209 \times 9)$ & MAE & $0.03447315$ & $0.08627122$ & $0.5221133$ & $0.00033116$ & $0.00098835$ \\
 & Pos Error & $0.02909847$ & $0.09582179$ & $0.68427341$ & $0.00044175$ & $0.00152845$ \\
 & Neg Error & $0.04264265$ & $0.03893273$ & $0.21990582$ & $0.0002631$ & $0.00086121$ \\  \hline
Daily\_Demand\_Forecasting\_Orders & RMSE & $0.03665407$ & $0.09586042$ & $1.08419387$ & $0.00294797$ & $0.00195928$ \\   
$(60 \times 12)$ & MAE & $0.02895825$ & $0.06542136$ & $0.67679833$ & $0.00109839$ & $0.00109826$ \\
 & Pos Error & $0.03217199$ & $0.11739298$ & $0.86489816$ & $0.00176665$ & $0.00180286$ \\
 & Neg Error & $0.01770995$ & $0.03737439$ & $0.30059867$ & $0.00043014$ & $0.00039365$ \\  \hline
gas\_furnace2 & RMSE & $0.03371765$ & $0.13730804$ & $0.66500792$ & $0.00003184$ & $0.00023341$ \\
$(293 \times 6)$ & MAE & $0.02870417$ & $0.6032496$ & $0.54300641$ & $0.00002163$ & $0.00014806$ \\
 & Pos Error & $0.03374757$ & $0.91916708$ & $0.63810399$ & $0.00020692$ & $0.00004956$ \\
 & Neg Error & $0.0255287$ & $0.07110268$ & $0.49128667$ & $0.00002272$ & $0.00015791$ \\   \hline
machine & RMSE & $0.04578493$ & $0.25929231$ & $1.42404944$ & $0.00636192$ & $0.00264685$ \\
$(209 \times 9)$ & MAE & $0.03507087$ & $0.00006935$ & $0.57988418$ & $0.00041585$ & $0.00191738$ \\
 & Pos Error & $0.03397687$ & $0.00006611$ & $0.94066641$ & $0.00056675$ & $0.00251774$ \\
 & Neg Error & $0.03684863$ & $0.02890076$ & $0.25190033$ & $0.00253411$ & $0.00143709$ \\  \hline
wpbc & RMSE & $0.00139195$ & $0.86978316$ & $0.95779329$ & $0.00023423$ & $0.00019688$ \\
$(194 \times 34)$ & MAE & $0.00133402$ & $0.71006265$ & $0.63237936$ & $0.00060265$ & $0.00009768$ \\
 & Pos Error & $0.00125402$ & $0.66320246$ & $1.31818329$ & $0.00075996$ & $0.00009044$ \\
 & Neg Error & $0.00219808$ & $0.45318878$ & $0.3771965$ & $0.00005219$ & $0.00010569$ \\  \hline
Average RMSE &  & $0.07662296$ & $0.28725999$ & $0.81334001$ & $\underline{0.0016327}$ & $\textbf{0.00084411}$ \\  \hline  
Average rank &  & $3.2$ & $4$ & $4.8$ & $1.7$ & $1.3$  \\  \hline
\multicolumn{7}{l}{$^{\dagger}$ represents the proposed model.} \\
\multicolumn{7}{l}{The top and second-best models in terms of RMSE are denoted by boldface and underline, respectively.}
\end{tabular}}
\end{table*}

\subsection{Experiments on Real World Regression Datasets}
In this subsection, we evaluate the performance of the proposed TRKM-R model using $10$ benchmark regression datasets from the UCI repository \cite{dua2017uci} against baseline SVR \cite{basak2007support}, TSVR \cite{peng2010tsvr}, TSVQR \cite{ye2024twin}, and RKM \cite{suykens2017deep} models. Table \ref{Average RMSE and average rank for UCI datasets} presents the detailed results of the proposed TRKM-R model and the existing models (SVR, TSVR, TSVQR, and RKM), evaluated using metrics such as RMSE, MAE, Pos Error, and Neg Error. RMSE, an important statistic whose lower values indicate better model performance, is the primary focus of the evaluation. For instance, on the Abalone and auto-original datasets, TRKM-R achieves remarkably low RMSE values, indicating its robustness in handling datasets with nonlinear and noisy relationships between features and target variables. Similarly, on the Airfoil\_Self\_Noise and bodyfat datasets, the model maintains strong predictive accuracy, effectively minimizing both positive and negative errors. For smaller datasets such as Auto-price and wpbc, the TRKM-R model demonstrates stable performance, yielding lower RMSE and MAE values compared to SVR and TSVR models, which tend to exhibit higher variance due to limited training samples. The model also performs favorably on the cpu\_perf and machine datasets, where complex feature interactions make regression challenging. TRKM-R efficiently captures these relationships, producing more accurate predictions than TSVQR and RKM. Moreover, in the Daily\_Demand\_Forecasting\_Orders and gas\_furnace2 datasets, TRKM-R successfully generalizes to temporal and dynamic data patterns, further emphasizing its flexibility in handling diverse regression scenarios. These consistent results across multiple datasets demonstrate that TRKM-R effectively balances model complexity and generalization, outperforming baseline approaches in both low- and high-dimensional settings. In $7$ of the $10$ datasets, the proposed TRKM-R model achieves the lowest RMSE values, indicating superior performance. For the remaining $3$ datasets, TRKM-R ranks second in terms of RMSE. This consistent performance across a wide range of datasets highlights the strength of the TRKM-R model. The effectiveness of the TRKM-R model is further confirmed by its average RMSE values. The average RMSE values for the existing SVR, TSVR, TSVQR, and RKM models are $0.07662296$, $0.28725999$, $0.81334001$, and $0.0016327$, respectively. In comparison, the proposed TRKM-R model achieves a superior average RMSE of $0.00084411$, outperforming the baseline models. To evaluate model performance accurately, it is important to rank each model separately for every dataset, rather than relying solely on the average RMSE values. Table \ref{Average RMSE and average rank for UCI datasets} displays the average ranking of the proposed TRKM-R model compared to the baseline models. Models are ranked based on RMSE, with the lowest RMSE receiving the highest rank. The average ranks of the TRKM-R model and the existing SVR, TSVR, TSVQR, and RKM models are $1.3$, $3.2$, $4$, $4.8$, and $1.7$, respectively. The rankings show the TRKM-R model's improved performance, decisively outperforming the baseline models.
We conducted the Friedman test \cite{demvsar2006statistical} and the Nemenyi post hoc test to further evaluate the efficacy of the proposed TRKM-R model. The significance of the performance variations between the models is statistically evaluated using the Friedman test. For $p=5$ and $N=10$, we obtained $\chi_F^2 = 35.44$ and $F_F = 69.947$. The $F_F$ statistic follows an $F$-distribution with degrees of freedom $(4, 36)$. According to the statistical $F$-distribution table, the critical value for $F_F(4,36)$ at a $5\%$ significance level is $2.6335$. Since the calculated $F_F$ value exceeds $2.6335$, the null hypothesis is rejected, indicating significant differences exist among the models. To identify significant differences in the pairwise comparisons between the models, the Nemenyi post-hoc test is used. We compute the $C.D.$ as $1.1364$, which means that for the average rankings in Table \ref{Average RMSE and average rank for UCI datasets} to be deemed statistically significant, there must be a minimum difference of $1.1364$ between them. The average rank distinctions between the TRKM-R model and the existing SVR, TSVR, TSVQR, and RKM models are $1.90$, $2.70$, $3.50$, and $0.40$, respectively. The proposed TRKM-R model is statistically better than other baseline models, except RKM, according to the Nemenyi post hoc test. The lower ranks of the proposed TRKM-R model indicate its stronger generalization capability compared to the existing RKM model. The combination of elevated average RMSE and consistent performance across multiple statistical tests provides strong evidence that the proposed TRKM-R model exceeds the performance of the existing baseline models in terms of generalization.

\subsection{Evaluation on  Brain Age Prediction}
In this subsection, we perform experiments on a brain age estimation dataset. The dataset comprises structural MRI scans from a total of 976 subjects, which were sourced from two prominent repositories: the OASIS dataset\footnote{\url{https://sites.wustl.edu/oasisbrains/}} and the IXI dataset\footnote{\url{https://brain-development.org/ixi-dataset/}}. The OASIS dataset provides MRI data for aging studies, while the IXI dataset includes diverse brain imaging data from different age groups. This comprehensive dataset enables robust evaluation and validation of the proposed TRKM-R model for predicting brain age. The dataset consists of $30$ patients diagnosed with Alzheimer's disease (AD), $876$ cognitively healthy (CH) individuals, and $70$ patients with mild cognitive impairment (MCI) \cite{ganaie2022brain}. For training the brain age predictor, $90\%$ of the cognitively healthy subjects were randomly selected, totaling $788$ individuals with a mean age of $47.40 \pm 19.69$ years. For model validation, we used the remaining cognitively healthy subjects, amounting to $88$ individuals with a mean age of $48.17 \pm 17.73$ years, along with the MCI patients ($70$ subjects with a mean age of $76.21 \pm 7.18$ years) and AD patients ($30$ subjects with a mean age of $78.03 \pm 6.91$ years).
\begin{figure*}[ht!]
\begin{center}
\begin{minipage}{.32\linewidth}
\centering
\subfloat[SVR]{\includegraphics[scale=0.33]{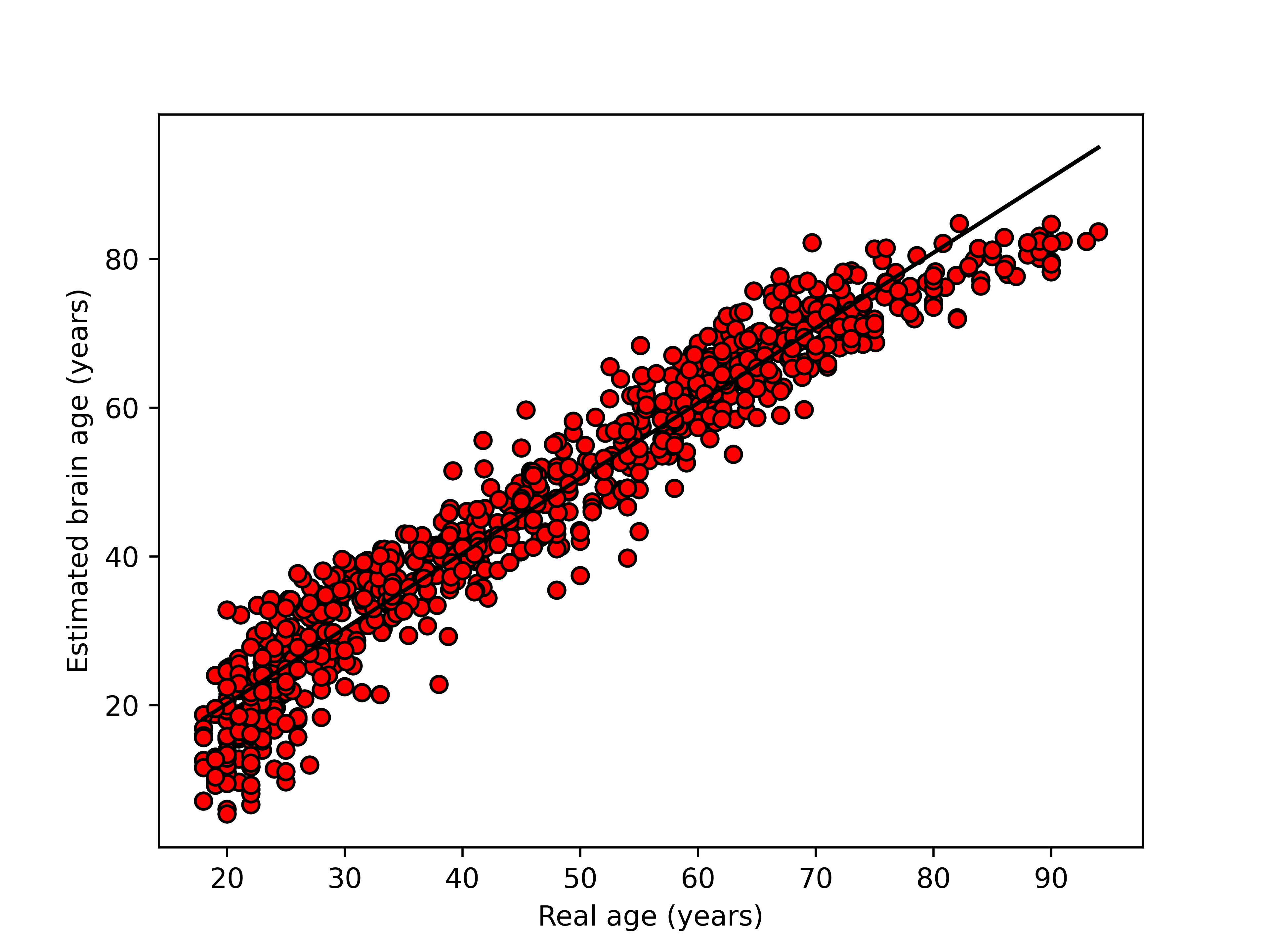}}
\end{minipage}
% \par\medskip
% \par\medskip
\begin{minipage}{.32\linewidth}
\centering
\subfloat[TSVR]{\includegraphics[scale=0.33]{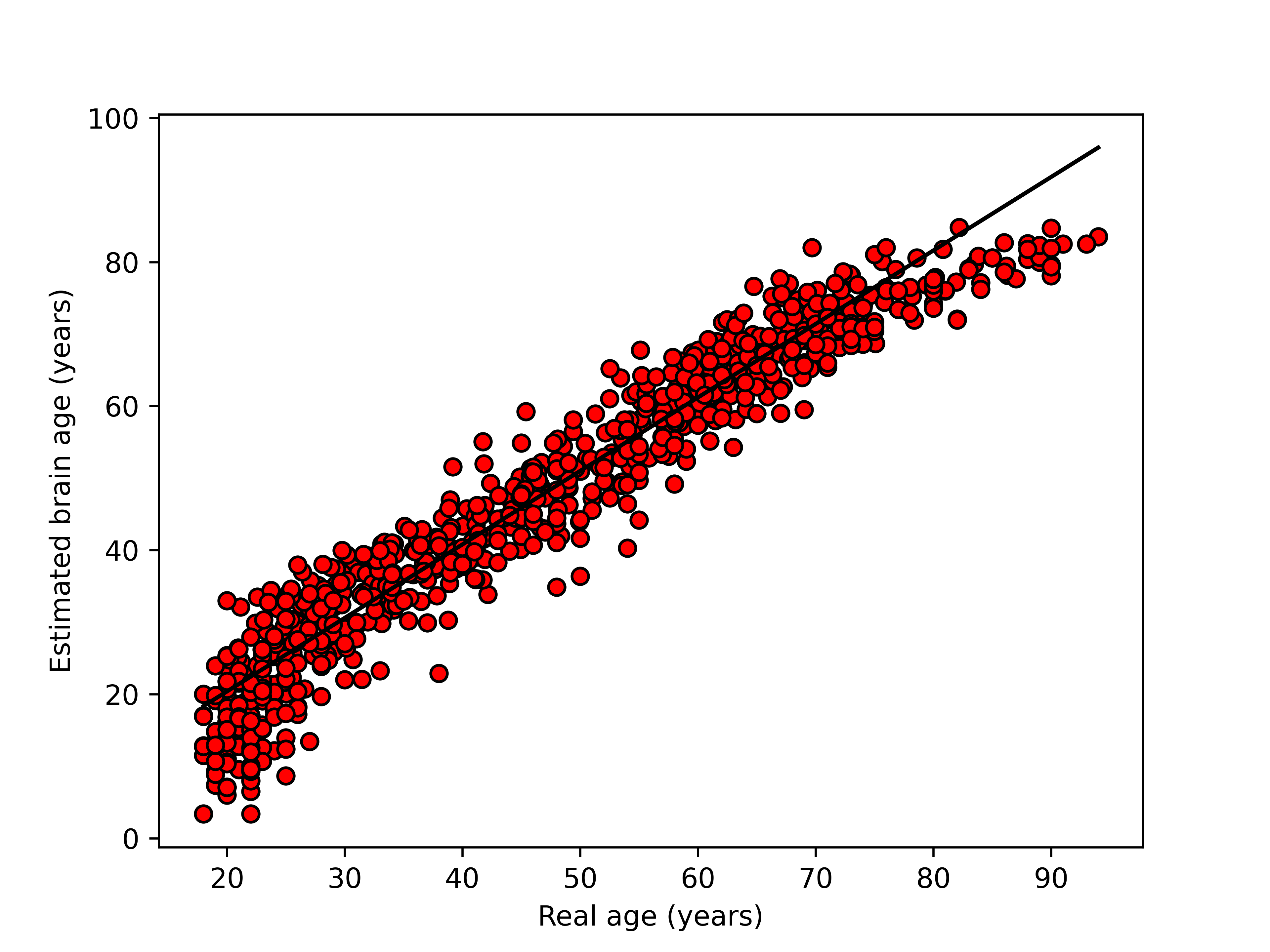}}
\end{minipage}
% \par\medskip
\begin{minipage}{.32\linewidth}
\centering
\subfloat[TSVQR]{\includegraphics[scale=0.33]{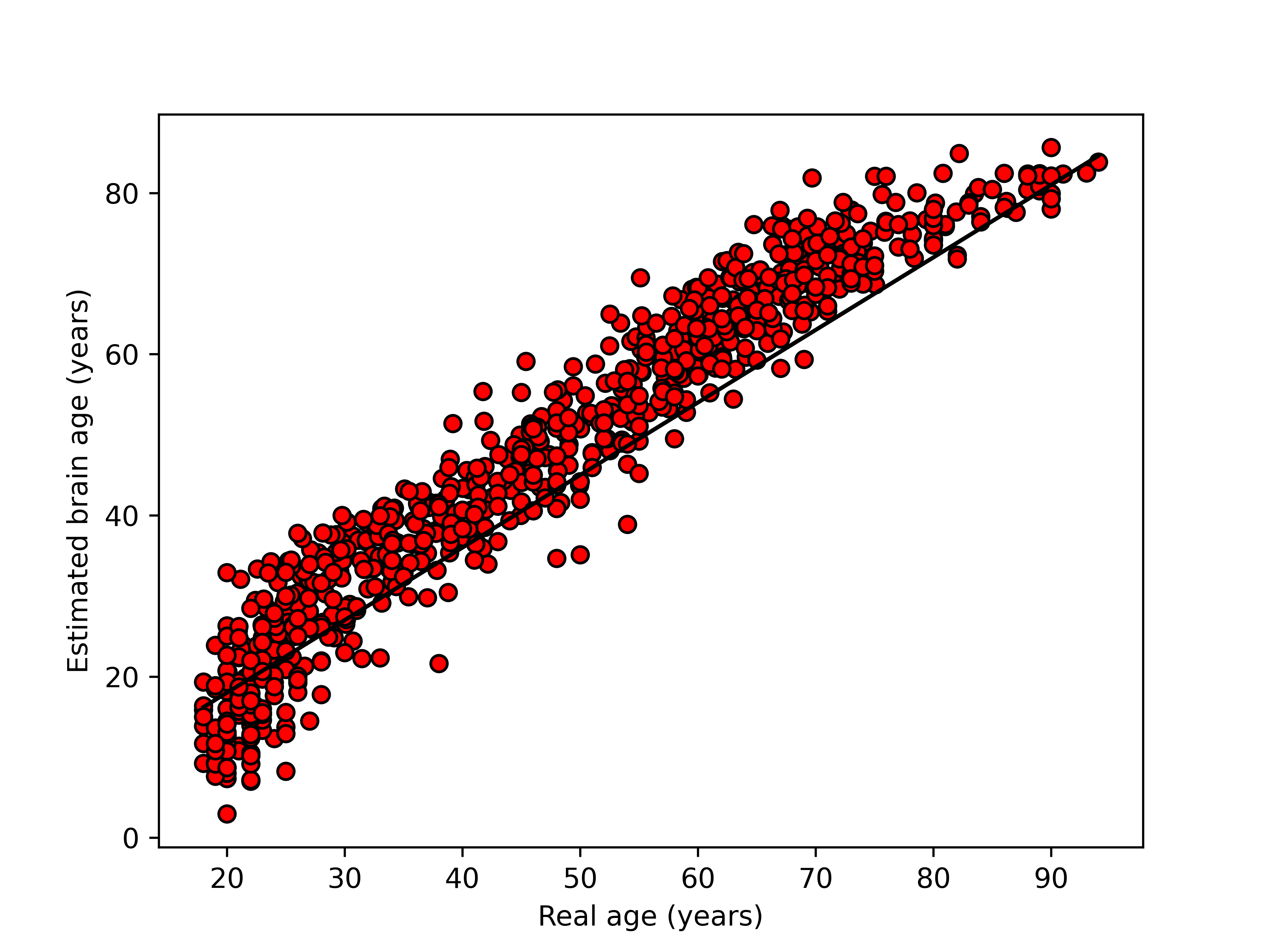}}
\end{minipage}
\end{center}
\par\medskip
\par\medskip
\begin{center}
\begin{minipage}{.32\linewidth}
\centering
\subfloat[RKM]{\includegraphics[scale=0.33]{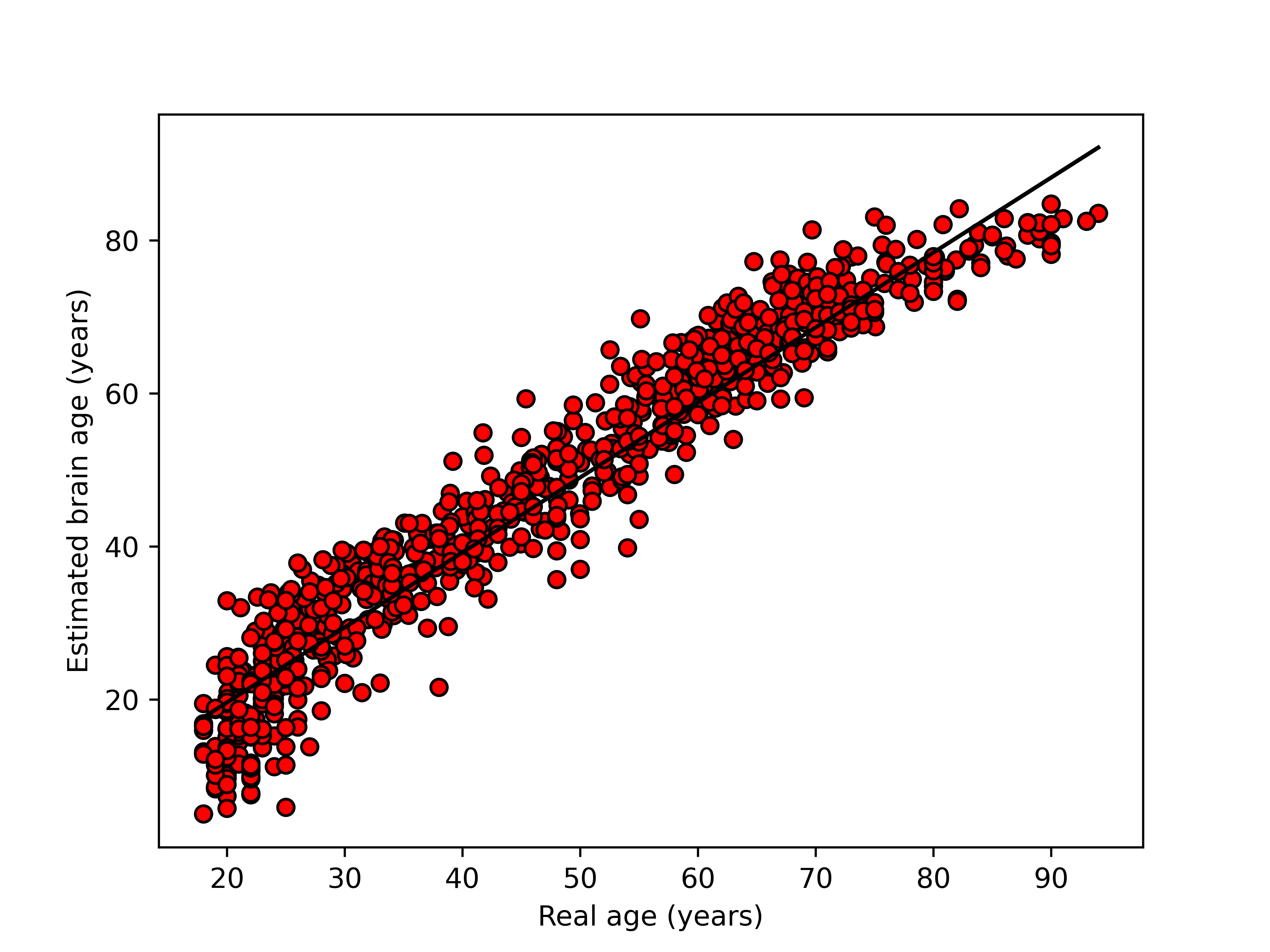}}
\end{minipage}
\begin{minipage}{.32\linewidth}
\centering
\subfloat[TRKM-R]{\includegraphics[scale=0.33]{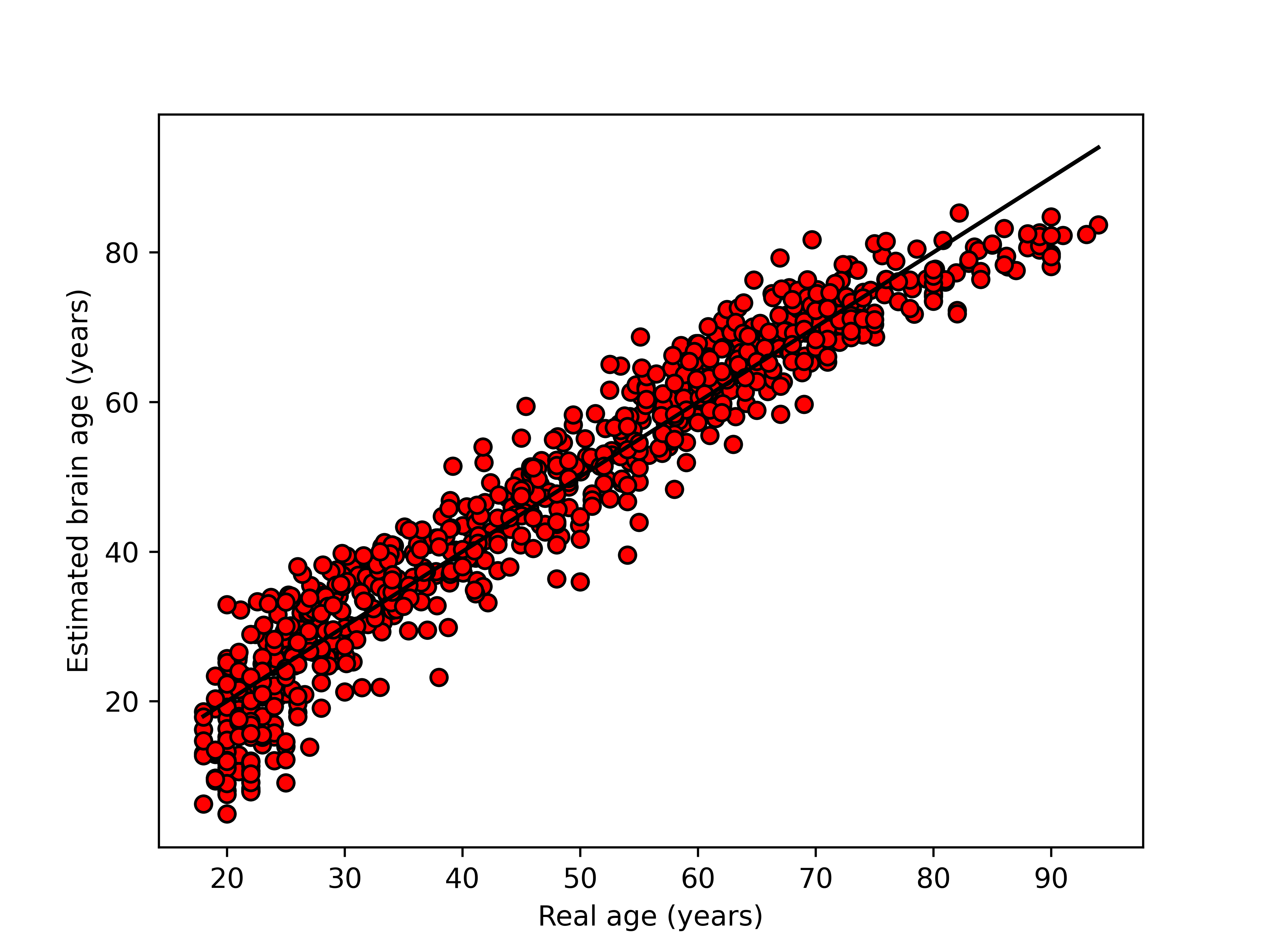}}
\end{minipage}
\end{center}
\caption{Real age versus estimated brain age on the training data using different prediction models, with the identity line represented by a black line.}
\label{The plot shows estimated brain age versus real age on the training data.}
\end{figure*}
The structural MRI scans were pre-processed using the CAT12 package\footnote{\url{http://dbm.neuo.uni-jena.de/}} and SPM12 software\footnote{\url{https://www.fil.ion.ucl.ac.uk/spm/}}. Initially, the MRI scans are divided into their fundamental components: white matter (WM), gray matter (GM), and cerebrospinal fluid (CSF). This paper specifically concentrates on the GM data. The GM images are registered to the Montreal Neurological Institute (MNI) template using a diffeomorphic registration algorithm, after which they are modulated to conform to the template. A Gaussian smoothing filter with a full-width half-maximum (FWHM) of $4$ millimetres is used to further process the smoothed GM pictures after they have been resampled to an isotropic spatial resolution of $8$ millimetres. As a result of this process, each subject yields approximately $3,700$ GM voxel values. To evaluate the prediction accuracy of the models, we compute several metrics: root mean square error (RMSE), mean absolute error (MAE), coefficient of determination (\( R^2 \)), brain-age delta (\( \Delta \)), and the $95\%$ confidence interval (CI). After training each model on the training set, it is then used to estimate brain age for the independent test sets.

\subsubsection{Experimental Results on the Training Set}

\begin{table*}[ht!]
\centering
    \caption{Comparison of the proposed TRKM-R model's performance against baseline models using training data of Brain Age estimation.}
    \label{Average RMSE and average rank for brain training}
    \resizebox{0.9\linewidth}{!}{
\begin{tabular}{lccccc}
\hline
Model $\rightarrow$ & SVR \cite{basak2007support} & TSVR \cite{peng2010tsvr} & TSVQR \cite{ye2024twin} & RKM \cite{suykens2017deep} & TRKM-R$^{\dagger}$ \\ \hline
MAE & $7.41$ & $6.83$ & $6.27$ & $4.79$ & $3.27$ \\
RMSE & $8.79$ & $7.11$ & $6.79$ & $5.33$ & $3.92$ \\
Mean brain age delta & $0$ & $0$ & $0$ & $0$ & $0$ \\
$95\%$ CI Values & $[-0.54,  0.54]$ & $[-0.52,  0.52]$ & $[-0.54,  0.54]$ & $[-0.44,  0.44]$ & $[-0.29,  0.29]$ \\
$R^2$ Score & $0.88$ & $0.88$ & $0.88$ & $0.92$ & $0.96$ \\ \hline
\multicolumn{6}{l}{$^{\dagger}$ represents the proposed model.}
\end{tabular}}
\end{table*}

Table \ref{Average RMSE and average rank for brain training} shows the performance of the proposed TRKM-R and the baseline models on the training set. The prediction accuracy shows that the SVR model has an MAE of $7.41$ years, followed by the TSVR, TSVQR, and RKM models are $6.83$, $6.27$, and $4.79$ years, and the proposed TRKM-R with $3.27$ years. RMSE values show a similar pattern, with the SVR model at $8.79$ years, the TSVR, TSVQR, and RKM models at $7.11$, $6.79$, and $5.33$ years, respectively, and the proposed TRKM model achieving $3.92$ years. As illustrated in Table \ref{Average RMSE and average rank for brain training}, the proposed TRKM-R model outperforms the existing SVR, TSVR, TSVQR, and RKM models. For all prediction models in the training set, the average brain age delta is zero. The training set's predicted outcomes for each model failed to demonstrate a statistically significant age dependence (\(P > 0.05\)). Fig. \ref{The plot shows estimated brain age versus real age on the training data.} illustrates the relationship between the actual age and estimated brain age for different prediction models in the training set. As shown in Fig. \ref{The plot shows estimated brain age versus real age on the training data.}, the proposed TRKM-R significantly outperformed the baseline models. The SVR, TSVR, and TSVQR models have similar prediction \( R^2 \) scores (\( R^2 = 0.88 \)), while the RKM model has an \( R^2 \) of 0.92. In contrast, the proposed TRKM model achieved a significantly higher \( R^2 \) of 0.96 on the same data.

\subsubsection{Experimental Results on Independent Test Sets}

\begin{table*}[htp]
\centering
    \caption{Comparison of the proposed TRKM-R model's performance against baseline models using testing Data (AD, CH, and MCI subjects).}
    \label{Average RMSE and average rank for brain datasets}
    \resizebox{0.9\linewidth}{!}{
\begin{tabular}{llccccc}
\hline
Dataset $\downarrow$ Model $\rightarrow$ &  & SVR \cite{basak2007support} & TSVR \cite{peng2010tsvr} & TSVQR \cite{ye2024twin} & RKM \cite{suykens2017deep} & TRKM-R$^{\dagger}$ \\ \hline
AD & MAE & $8.38$ & $8.82$ & $7.85$ & $9.67$ & $7.67$ \\
 & RMSE & $12.64$ & $11.46$ & $11.96$ & $8.89$ & $8.49$ \\
 & Mean brain age delta & $7.64$ & $5.01$ & $6.96$ & $6.55$ & $5.75$ \\
 & $95\%$ CI Values & $[8.22,  13.06]$ & $[8.26,  12.77]$ & $[6.54,  13.38]$ & $[8.85,  14.25]$ & $[3.85,  9.25]$ \\
 & $R^2$ Score & $0.24$ & $0.3$ & $0.12$ & $0.21$ & $0.31$ \\ \hline
CH & MAE & $14.77$ & $9.33$ & $12.45$ & $7.84$ & $5.32$ \\
 & RMSE & $7.65$ & $5.94$ & $8.81$ & $6.12$ & $5.74$ \\
 & Mean brain age delta & $0.78$ & $1.03$ & $-3.9$ & $1.18$ & $1.18$ \\
 & $95\%$ CI Values & $[-1.54,  2.98]$ & $[-1.58,  1.51]$ & $[-1.15,  1.66]$ & $[-2.89,  1.35]$ & $[-2.72,  0.36]$ \\
 & $R^2$ Score & $0.87$ & $0.84$ & $0.97$ & $0.59$ & $0.84$ \\ \hline
MCI & MAE & $8.82$ & $8.6$ & $7.84$ & $8.98$ & $7.31$ \\
 & RMSE & $9.69$ & $11.84$ & $14.14$ & $9.03$ & $7.03$ \\
 & Mean brain age delta & $4.82$ & $4.31$ & $4.14$ & $4.97$ & $3.97$ \\
 & $95\%$ CI Values & $[-5.53,  3.11]$ & $[-5.75,  2.38]$ & $[-5.85,  2.42]$ & $[-5.13,  2.08]$ & $[-5.13,  2.08]$ \\
 & $R^2$ Score & $0.32$ & $0.22$ & $0.13$ & $0.23$ & $0.97$ \\ \hline
 \multicolumn{7}{l}{$^{\dagger}$ represents the proposed model.}
\end{tabular}}
\end{table*}

\begin{figure*}[ht!]
\begin{center}
\begin{minipage}{.30\linewidth}
\centering
\subfloat[SVR]{\includegraphics[scale=0.17]{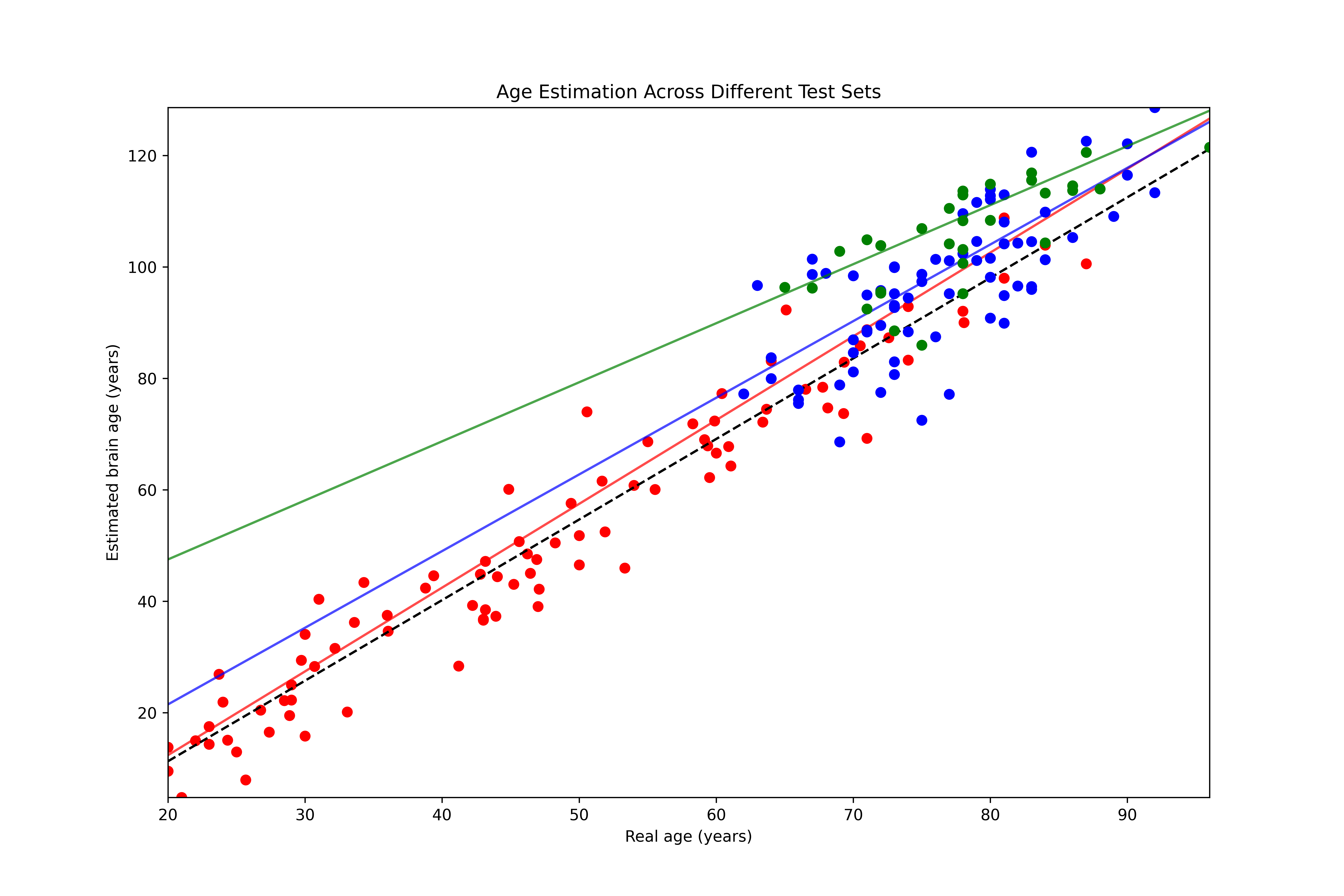}}
\end{minipage}
% \par\medskip
% \par\medskip
\begin{minipage}{.30\linewidth}
\centering
\subfloat[TSVR]{\includegraphics[scale=0.17]{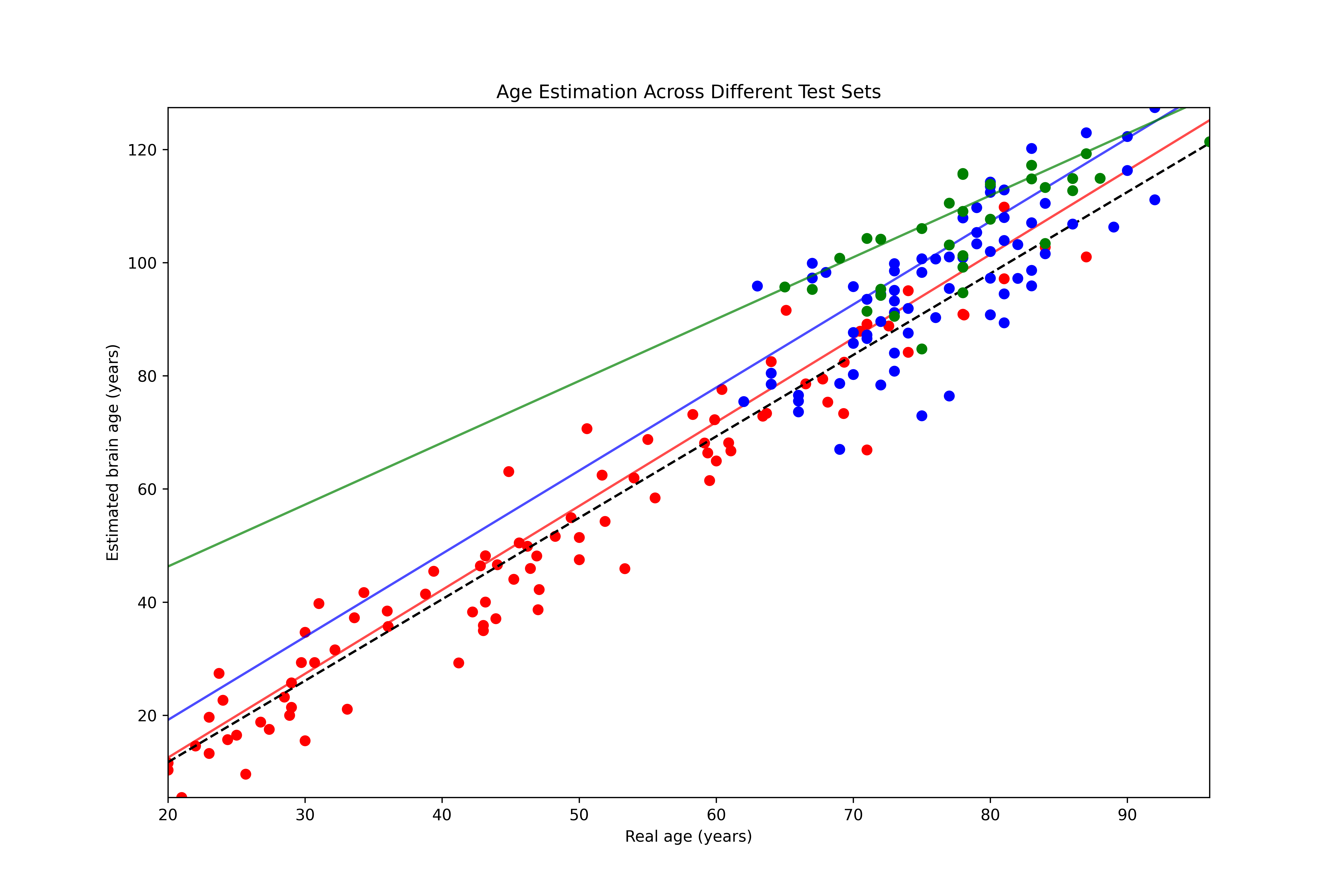}}
\end{minipage}
% \par\medskip
\begin{minipage}{.30\linewidth}
\centering
\subfloat[TSVQR]{\includegraphics[scale=0.17]{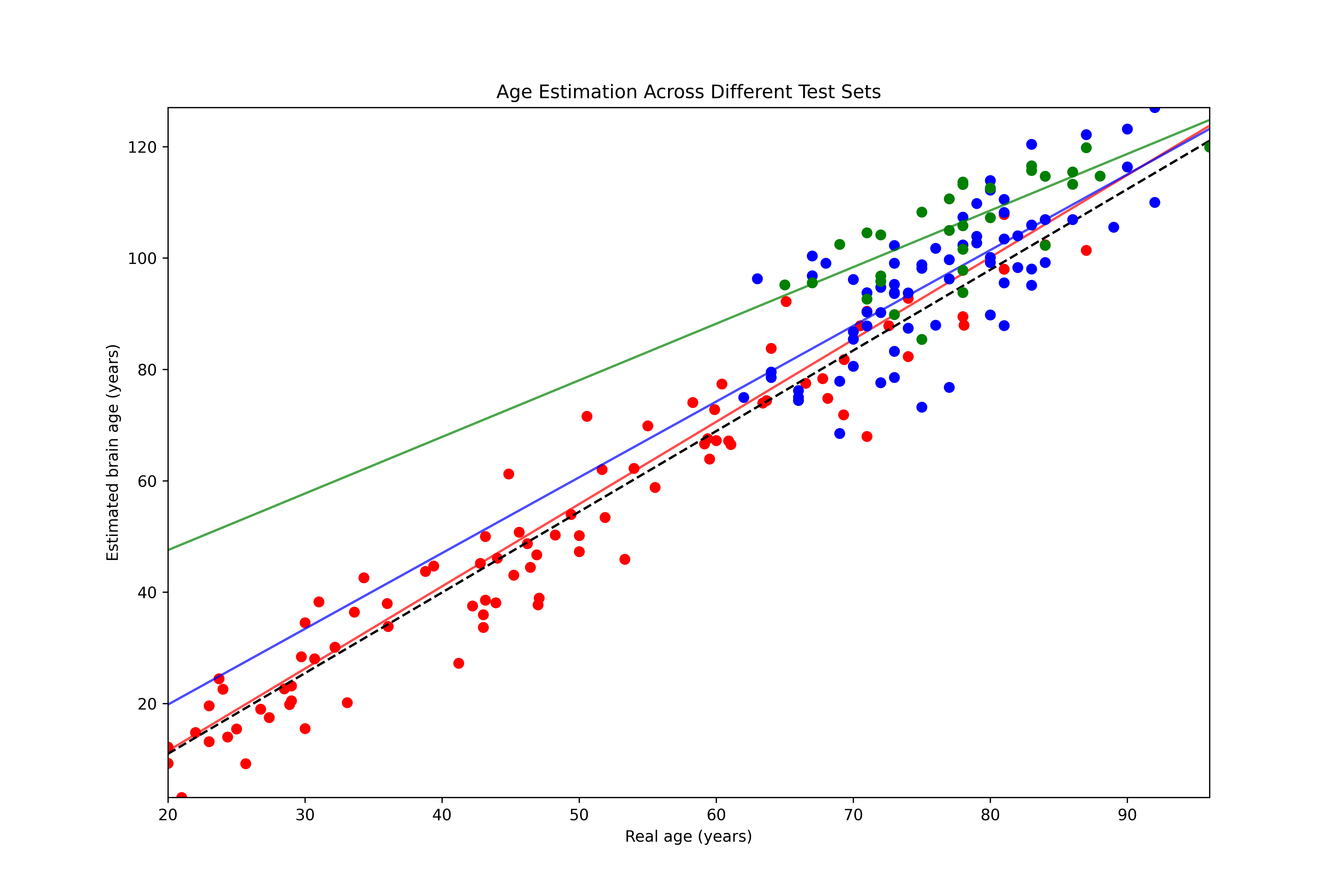}}
\end{minipage}
\end{center}
\par\medskip
\par\medskip
\begin{center}
\begin{minipage}{.30\linewidth}
\centering
\subfloat[RKM]{\includegraphics[scale=0.17]{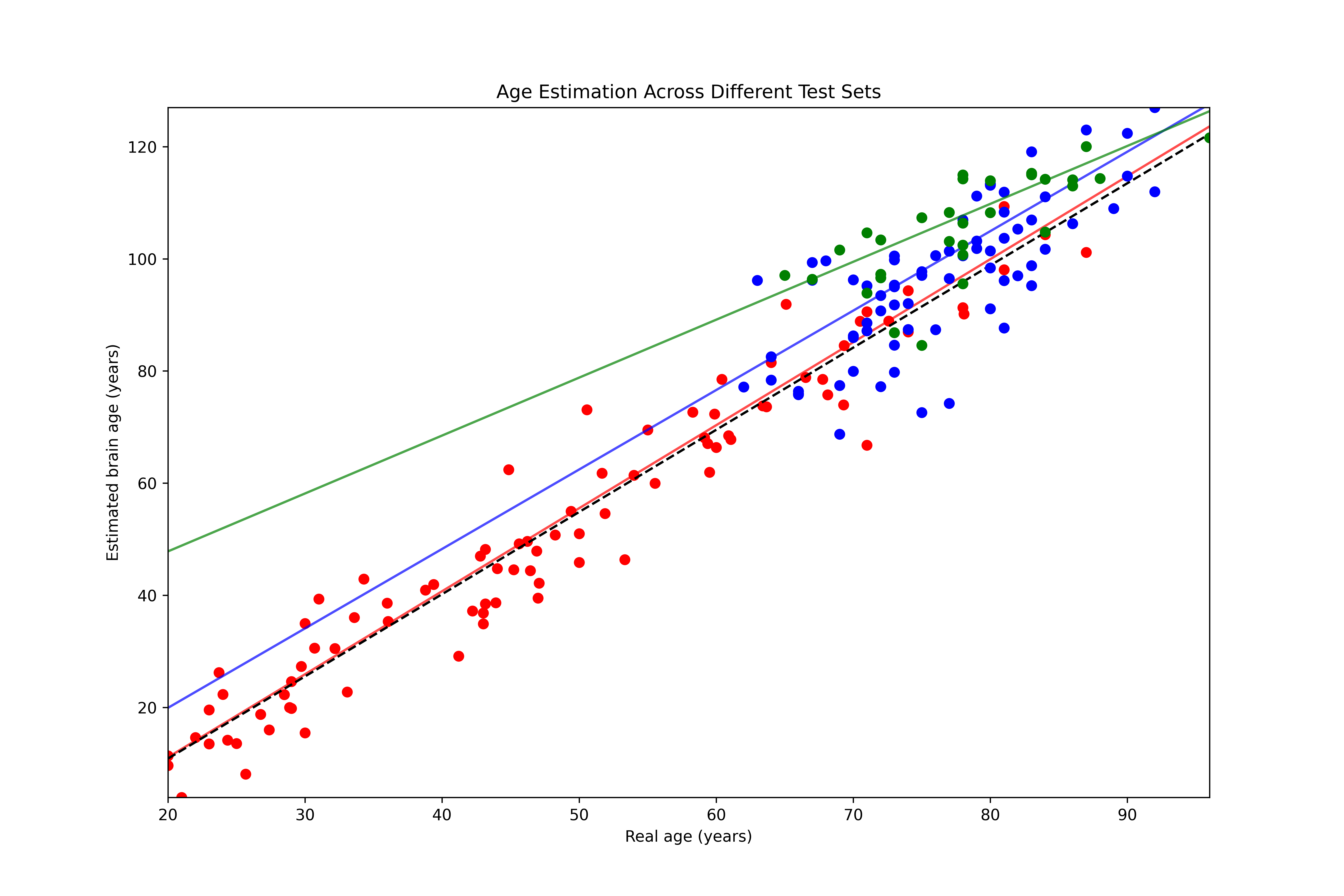}}
\end{minipage}
\begin{minipage}{.30\linewidth}
\centering
\subfloat[TRKM-R]{\includegraphics[scale=0.17]{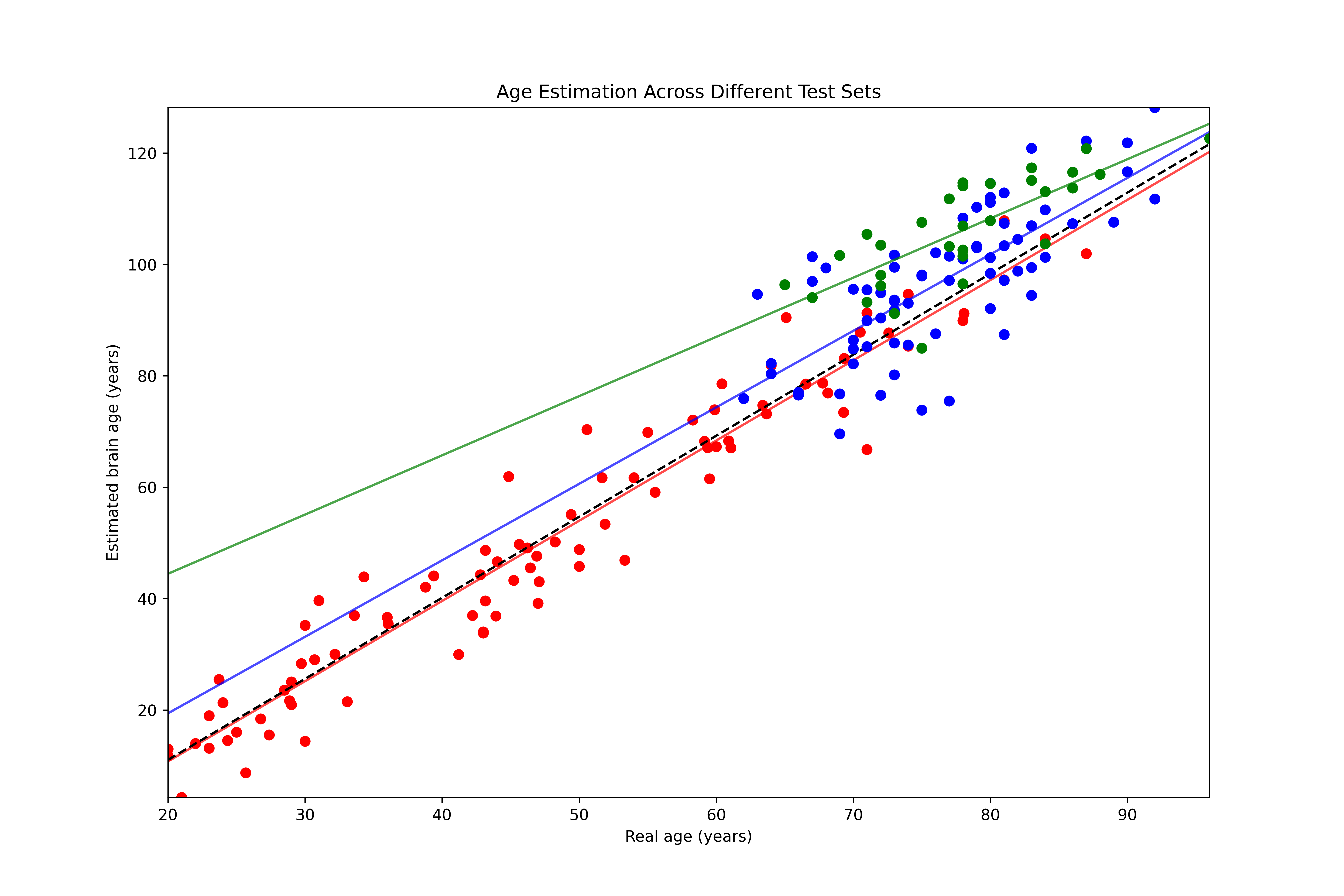}}
\end{minipage}
\end{center}
\caption{Estimated actual age versus brain age on the independent test sets features red markers and regression lines for CH subjects, blue markers, and regression lines for MCI subjects, and green markers and regression lines for AD subjects. An identity line, shown in black, serves as a reference for comparison across different prediction models.}
\label{The plot shows estimated brain age versus real age on the independent test}
\end{figure*}

Table \ref{Average RMSE and average rank for brain datasets} shows the performance of the TRKM-R model against the baseline models on independent test sets. TRKM-R model attained an MAE of $5.32$ years on cognitively healthy subjects, significantly outperforming the SVR, TSVR, TSVQR, and RKM models, which had $MAEs$ of $14.77$, $9.33$, $12.45$, and $7.84$ years, respectively. In the independent group of cognitively healthy individuals, the mean brain age delta was approximately zero across all prediction models. The analysis of independent test sets reveals that the predicted brain age for cognitively healthy individuals did not exhibit significant age dependency across different models (\( P > 0.05 \)), consistent with the training set results. In contrast, both AD and MCI groups showed a positive brain age delta across all models, indicating that their predicted brain age was higher than their actual age. For the MCI group, the mean brain age delta values ranged from \( \Delta = 3.97 \) years (TRKM-R) to \( \Delta = 4.97 \) years (RKM), while for the AD group, the values ranged from \( \Delta = 5.75 \) years (TRKM-R) to \( \Delta = 7.64 \) years (SVR). These results suggest that the TRKM-R model offers a lower brain age delta compared to most baseline models, reflecting better performance in brain age estimation. Fig. \ref{The plot shows estimated brain age versus real age on the independent test} illustrates the correlation between predicted and actual brain ages, where different markers and regression lines for CH, MCI, and AD subjects show how well the models estimate brain age across these conditions.

All prediction models revealed a positive brain age delta for both MCI and AD subjects, as shown in Table \ref{Average RMSE and average rank for brain datasets}. This positive delta indicates that these individuals, on average, show an accelerated brain aging process compared to age-matched healthy controls. A higher brain age delta correlates with increased brain atrophy and deterioration in cognitive functions such as memory, attention, and problem-solving skills in individuals with AD \cite{beheshti2018association}. This implies that MCI and AD patients generally experience more substantial brain atrophy than their healthy peers.

Specifically, individuals with AD exhibit a greater brain age delta compared to those with MCI, suggesting that AD is associated with more pronounced brain atrophy. The greater delta observed in AD patients relative to MCI patients points to a more rapid progression of brain aging in AD \cite{beheshti2018association}. Furthermore, as illustrated in Fig. \ref{The plot shows estimated brain age versus real age on the independent test}, younger patients with MCI or AD show a larger brain age discrepancy than older patients within the same diagnostic category. This finding aligns with other studies suggesting that early-onset AD patients typically exhibit a larger brain age gap compared to those with late-onset AD. The proposed TRKM-R model achieved an MAE of $5.32$ years when applied to cognitively healthy subjects \cite{beheshti2021disappearing}.

\section{Conclusion}
\label{Conclusion}
In this paper, we proposed a novel twin restricted kernel machine (TRKM) model for classification and regression. The TRKM model effectively tackles the challenges related to generalization in RKMs, especially when working with unevenly distributed or complexly clustered data. By integrating the strengths of twin models and leveraging the conjugate feature duality based on the Fenchel-Young inequality, TRKM offers a robust and efficient framework for both classification and regression tasks. To evaluate the effectiveness, scalability, and efficiency of the proposed TRKM model, we conducted a series of rigorous experiments and performed comprehensive statistical analyses. We evaluated the proposed TRKM model using benchmark datasets from UCI and KEEL repositories and compared its performance against five state-of-the-art models for classification and regression. Here are the key findings: $(i)$ The results emphasize the outstanding performance of the TRKM model, which emerged as the top-performing model, achieving an average accuracy improvement of up to $0.42\%$ over the second-highest baseline model. $(ii)$ We conducted experiments on a regression task, and our proposed model achieved lower RMSE and MAE values, demonstrating its superior performance compared to the baseline models. $(iii)$ We evaluated the models on Brain Age Prediction datasets, where our proposed models demonstrated superior performance, showcasing their scalability and efficiency in handling this dataset effectively. $(iv)$ Statistical analyses—encompassing ranking, the Friedman test, the Nemenyi post hoc test, and the win-tie-loss sign test—demonstrate that our proposed model significantly outperforms the baseline models in terms of generalization performance. 

Our proposed TRKM model has demonstrated exceptional performance; however, it is currently restricted to handling vector-based inputs only. Future research should focus on extending the TRKM framework to accommodate matrix-structured data, exploring methods to reduce computational complexity, and broadening its applicability to more diverse datasets. Another potential direction is the integration of more advanced kernel functions or the development of adaptive kernel strategies that can dynamically adjust to the underlying data distribution, enhancing performance in highly heterogeneous datasets.
\section*{Acknowledgement}
This study receives support from the Science and Engineering Research Board (SERB) through the Mathematical Research Impact-Centric Support (MATRICS) scheme, Grant No. MTR/2021/000787. The authors gratefully acknowledge the invaluable support provided by the Indian Institute of Technology Indore.
\bibliography{refs.bib}
\bibliographystyle{unsrtnat}
\end{document}